\documentclass{article}

\usepackage{arxiv}
\usepackage{graphicx}%
\usepackage{multirow}%
\usepackage{amsmath,amssymb,amsfonts}%
\usepackage{caption}
\usepackage{amsthm}%
\usepackage{xspace}
\usepackage{xstring}
\usepackage{mathrsfs}%
\usepackage[title]{appendix}%
\usepackage{xcolor}%

\usepackage{hyperref}
\usepackage{natbib}
\usepackage{textcomp}%
\usepackage{manyfoot}%
\usepackage{xcolor,xstring,etoolbox}
\usepackage{booktabs}%
\usepackage{algorithm}%
\usepackage{algorithmicx}%
\usepackage{listings}%
\usepackage{geometry}
\usepackage{subcaption}
\usepackage{lmodern}
\usepackage{tabularx}
\usepackage[shortlabels]{enumitem}
\usepackage{longtable,lscape}
\usepackage{xcolor}

\theoremstyle{plain}
\usepackage{framed} 
\usepackage{lipsum} 

\theoremstyle{remark}

\theoremstyle{definition}
\usepackage[most]{tcolorbox}
\usepackage{listings} 
\usepackage{beramono} 

\usepackage{algpseudocode}
\algrenewcommand\algorithmicrequire{\textbf{Input:}}   
\algrenewcommand\algorithmicensure{\textbf{Output:}}   

\newtcolorbox{promptbox}[1][]{
    breakable,
    colback=blue!5!white,
    colframe=blue!40!black,
    fonttitle=\bfseries,
    coltitle=blue!40!black,
    colbacktitle=blue!10!white,
    sharp corners,
    boxrule=0.5pt,
    left=6pt, right=6pt, top=6pt, bottom=6pt,
    title={},
    #1
}

\definecolor{inputhighlight}{rgb}{0.2, 0.6, 0.2} 
\definecolor{instructionhighlight}{rgb}{0.8, 0.2, 0.2} 
\definecolor{strategyhighlight}{rgb}{0.2, 0.2, 0.8} 

\newcommand\MarkRedBibItems[1]{\gdef\RedBibKeys{,#1,}}
\MarkRedBibItems{lewis2020retrieval,edge2024local,wu_autogen_2023,chan2024chateval,wang2025mixtureofagents}

\makeatletter
\newif\if@lastred
\AtBeginEnvironment{thebibliography}{\@lastredfalse}
\AtEndEnvironment{thebibliography}{\if@lastred\color{black}\fi}
\pretocmd{\@bibitem}{%
  \if@lastred\color{black}\par\@lastredfalse\fi
  \IfSubStr{\RedBibKeys}{,#1,}{\color{black}\global\@lastredtrue}{}%
}{}{}
\pretocmd{\@lbibitem}{%
  \if@lastred\color{black}\par\@lastredfalse\fi
  \IfSubStr{\RedBibKeys}{,#2,}{\color{black}\global\@lastredtrue}{}%
}{}{}
\makeatother

\hypersetup{
  hidelinks,
  pdfsubject={Accepted by Neural Computing and Applications. Version of Record: https://doi.org/10.1007/s00521-026-12249-y}
}

\AtBeginDocument{%
  \renewcommand{\textcolor}[2]{#2}%
  \renewcommand{\color}[1]{}%
}

\begin{document}

\title{A Training-Free Mixture-of-Agents Framework for Multi-Document Summarization using LLMs and Knowledge Graphs}

\author{Cuong Vuong Tuan\textsuperscript{1}, Trang Mai Xuan\textsuperscript{1}\thanks{Corresponding author: trang.maixuan@phenikaa-uni.edu.vn}, Tien-Cuong Nguyen\textsuperscript{2}, Vu-Duc Ngo\textsuperscript{3}, Thien Van Luong\textsuperscript{4} \\
\textsuperscript{1}Faculty of Artificial Intelligence and Data Science, Phenikaa University, Duong Noi, Ha Noi, Viet Nam \\
\textsuperscript{2}VNPT AI, VNPT Group, Ha Noi, Viet Nam \\
\textsuperscript{3}MobiFone Research and Development Center, MobiFone Corporation, Ha Noi, Viet Nam \\
\textsuperscript{4}Business AI Lab, Faculty of Data Science and Artificial Intelligence, National Economics University, College of Technology, Ha Noi, Viet Nam \\
\texttt{cuong.vuongtuan@phenikaa-uni.edu.vn, trang.maixuan@phenikaa-uni.edu.vn, nguyentiencuong@vnpt.vn, duc.ngo@mobifone.vn, thienlv@neu.edu.vn}
}
\date{}

\maketitle
\begin{center}
\textbf{Comments:} The Version of Record of this article is published in \textit{Neural Computing and Applications}, and is available online at \href{https://doi.org/10.1007/s00521-026-12249-y}{https://doi.org/10.1007/s00521-026-12249-y}.
\end{center}

\begin{abstract}
Multi-Document Summarization (MDS) play a critical role in distilling essential information from collections of textual data. Existing approaches often struggle to capture complex inter-document relationships, rely heavily on large amounts of labeled data for supervised training, or exhibit limited generalization across domains and languages. To address these limitations, we present a training-free mixture-of-agents framework for MDS that leverages the complementary strengths of large language models (LLMs) and knowledge graphs. Our approach decomposes summarization into specialized agent tasks: extractive selection, knowledge-aware abstraction, and iterative refinement, each operating without task-specific fine-tuning. 
\textcolor{black}{We unify their outputs using a multi-perspective consistency mechanism guided by LLMs.} Experiments across four datasets in English and Vietnamese demonstrate state-of-the-art or competitive performance, validating the effectiveness and adaptability of our modular design.
\end{abstract}



\section{Introduction}\label{sec:intro}

Rapidly expanding digital information necessitates MDS for synthesizing knowledge from diverse sources, making it a critical task in natural language processing \citep{mridha2021survey, ma2022multi}. Early extractive methods \citep{erkan_lexrank_2004, mihalcea_textrank_2004} and rule-based abstractive systems \citep{li_hybrid_2017} often produce outputs that lack semantic coherence and fluency. The advent of deep learning, especially sequence-to-sequence architectures \citep{sutskever_seq2seq_2014} and subsequently transformer-based LLMs \citep{vaswani2017attention, wu2025survey}, has significantly improved
\textcolor{black}{the quality of abstractive summarization} \citep{zhang_pegasus_2020, xiao_primera_2022}. However, leading supervised models require large task-specific labeled datasets \citep{fabbri_multi-news_2019, lu_multi-xscience_2020}, limiting their applicability in low-resource domains or under-resourced languages such as Vietnamese \citep{tran2020vims}.

\textcolor{black}{Recent advances in LLMs have opened new opportunities for MDS through strong zero-shot and few-shot capabilities \citep{brown2020language, wei2022emergent}. However, context length limitations remain a significant barrier. Moreover, The et al. \citep{le2025latvis} conducted specific fine-tuning on Vietnamese datasets, achieving promising results. Nevertheless, these approaches face challenges including limited testing on diverse English datasets and requirements for extensive data collection and labeling. Furthermore, current training-free methods that use pre-trained LLMs encounter specific difficulties in learning complex inter-document relationships and handling large semantic contexts \citep{shakil2024abstractive}. Therefore, specialized frameworks are needed that can use existing LLM knowledge without requiring task-specific data collection or fine-tuning, while addressing coordination challenges across multiple documents.}

This paper introduces the \textbf{Mixture of Agents (MoA)} framework, a novel architectural approach for MDS that leverages pre-trained LLMs within a collaborative multi-agent system \textbf{without task-specific fine-tuning}. The MoA architecture decomposes the complex MDS task into complementary sub-problems managed by specialized agents as follows:
\begin{enumerate}[(1)]
    \item \textbf{Extractor agent}: Identifies salient factual sentences using a robust scoring model.
    \item \textbf{KGSum agent}: Our most sophisticated agent, which constructs a knowledge graph (KG) to model entities and their relationships explicitly. It then generates a KG-based summary by identifying thematic communities and detecting contrastive information within the graph, offering a deep, structured perspective.
    \item \textbf{Abstractor agent}: Produces a fluent, coherent abstractive summary directly from the source documents.
\end{enumerate}
The outputs from these parallel agents are orchestrated by our novel \textbf{Adaptive Multi-Perspective Fusion (AMF)} mechanism. AMF assesses agent-generated metadata to dynamically select the optimal strategy for synthesizing a final summary, robustly fusing extractive, KG-based, and abstractive perspectives.

Our primary contributions are as follows:
\begin{itemize}
    \item \textcolor{black}{We propose MoA, a novel architectural framework for training-free multi-document summarization that coordinates three specialized agents: Extractor, KGSum and Abstractor. The main innovation is the KGSum agent that constructs knowledge graphs to explicitly model complex inter-document relationships and contradictions, enabling systematic learning of intricate connections between multiple documents.}

    \item We introduce the Adaptive Multi-Perspective Fusion (AMF) mechanism that uses agent-generated metadata to dynamically determine optimal fusion strategies, allowing MoA to adaptively exploit agent-specific strengths based on document characteristics.
    
    \item We demonstrate MoA's effectiveness through comprehensive evaluations using various LLMs on benchmark datasets across English and Vietnamese, achieving state-of-the-art performance and validating our architectural approach for learning complex inter-document relationships.
\end{itemize}
The remainder of the paper is organized as follows: Section \ref{sec:related-work} discusses related work in MDS. Section \ref{sec:methodology} details the MoA framework. Section \ref{sec:experiments} presents the experimental setup, results, and analysis. Section \ref{sec:conclusion} concludes and suggests future directions.


\section{Related Work}\label{sec:related-work}

\subsection{Training-free Approaches for MDS}\label{subsec:training_free_approaches}

Early efforts in this paradigm were dominated by traditional unsupervised extractive techniques. For example, maximal marginal relevance, which selects relevant yet diverse sentences, was proposed in \citep{carbonell_mmr_1998}, while centroid-based approaches that identify sentences central to a document cluster's theme were introduced in \citep{radev_centroid_2004}. Additionally, graph-based algorithms such as  LexRank \citep{erkan_lexrank_2004} and TextRank \citep{mihalcea_textrank_2004} rely on inter-sentence relationships to rank sentences by centrality. \textcolor{red}{Early hybrid extractive-abstractive methods combined extractive ranking with abstractive generation \citep{li_hybrid_2017}.}

More recent unsupervised approaches have incorporated deep learning representations without task-specific fine-tuning. Examples include using pre-trained model embeddings within graph-based frameworks \citep{jia_graph_unsup_2020} or leveraging topic modeling with transformer representations, as seen in GLIMMER-TTR \citep{estape_glimmer_2021}. Unsupervised graph-based methods were also explored for languages such as Vietnamese \citep{nguyen_graph_hybrid_2020}.

The advent of LLMs has further expanded training-free MDS possibilities, primarily through zero-shot or few-shot prompting \citep{shakil2024abstractive}.
\textcolor{black}{Li et al. \citep{li_compress_mds_2023} focused on strategies to compress or select information before presenting concatenated documents to an LLM due to context window limitations.} While these LLM-driven prompting approaches offer flexibility and reduce the need for specific MDS training datasets, directly prompting a single LLM for complex MDS can struggle with information reconciliation and lack coherence across numerous documents.

\subsection{Full-training approaches for MDS}\label{subsec:full_training_approaches}

Contrasting with training-free methods, full training approaches involve extensive supervised learning on large-scale, task-specific MDS datasets. The advent of deep learning, particularly sequence-to-sequence models \citep{sutskever_seq2seq_2014}, significantly advanced abstractive summarization. For example, Pointer-Generator Networks \citep{see_pointer_generator_2017} was proposed to improve the factual consistency.

However, the paradigm shifted decisively with PLMs fine-tuned specifically for MDS. Architectures designed for long sequences (e.g., Longformer \citep{beltagy_longformer_2020}) were adapted. End-to-end supervised training on large MDS datasets, such as Multi-News \citep{fabbri_multi-news_2019} and Multi-XScience \citep{lu_multi-xscience_2020} for English, or VN-MDS and ViMs \citep{tran2020vims} for Vietnamese, led to powerful models, such as PEGASUS \citep{zhang_pegasus_2020} (gap-sentence objective), PRIMERA \citep{xiao_primera_2022} (entity-masking), and Centrum \citep{liu_centrum_2022} (discourse coherence). Some research works also explored fine-tuning general-purpose LLMs on MDS tasks, which falls under the full training paradigm due to their reliance on task-specific labeled data.

These models typically achieve SOTA ROUGE scores \citep{lin_rouge_2004} on in-domain data but are fundamentally dependent on such labeled datasets. This restricts their applicability in low-resource scenarios, novel domains, or for languages whose large specific MDS corpora are unavailable. MoA, as a training-free approach, aims to circumvent this data dependency entirely.

\subsection{Agent-Based Systems}\label{subsec:agent_based_systems}

\textcolor{black}{The emergence of knowledge-enhanced retrieval systems has shown promise for various AI tasks. Retrieval-Augmented Generation (RAG) \citep{lewis2020retrieval} represents a major advance in knowledge-intensive tasks, showing strong results in question-answering by combining retrieval with text generation. GraphRAG \citep{edge2024local} extends this approach using graph structures to capture relationships and enable better reasoning. However, these query-focused methods may not suit the multi-perspective analysis needed for comprehensive document summarization.}

\textcolor{black}{Recently, multi-agent systems powered by LLMs have gained significant attention for complex problem-solving and collaboration \citep{park_generative_agents_2023, wang_agent_survey_2023}. AutoGen \citep{wu_autogen_2023} enables multiple agents to converse and complete tasks through flexible interaction patterns. Similarly, ChatEval \citep{chan2024chateval} uses multi-agent debates where LLMs collaborate as evaluators to enhance task performance. Moreover, the Mixture-of-Agents approach \citep{wang2025mixtureofagents} constructs layered architectures where each agent leverages outputs from previous layer agents as auxiliary information to enhance response quality. While these frameworks excel in their respective domains, MDS presents unique challenges that benefit from parallel processing of diverse summarization perspectives rather than iterative debate refinement.} 

\textcolor{black}{To the best of our knowledge, this work presents the first attempt at developing a structured multi-agent MoA architecture specifically for MDS. Our key contribution lies in integrating AMF mechanisms within the MoA paradigm for training-free MDS, differentiating our approach from monolithic LLM prompting and traditional MDS techniques.}


\begin{figure}[ht]
  \centering
	\includegraphics[width=1\textwidth]{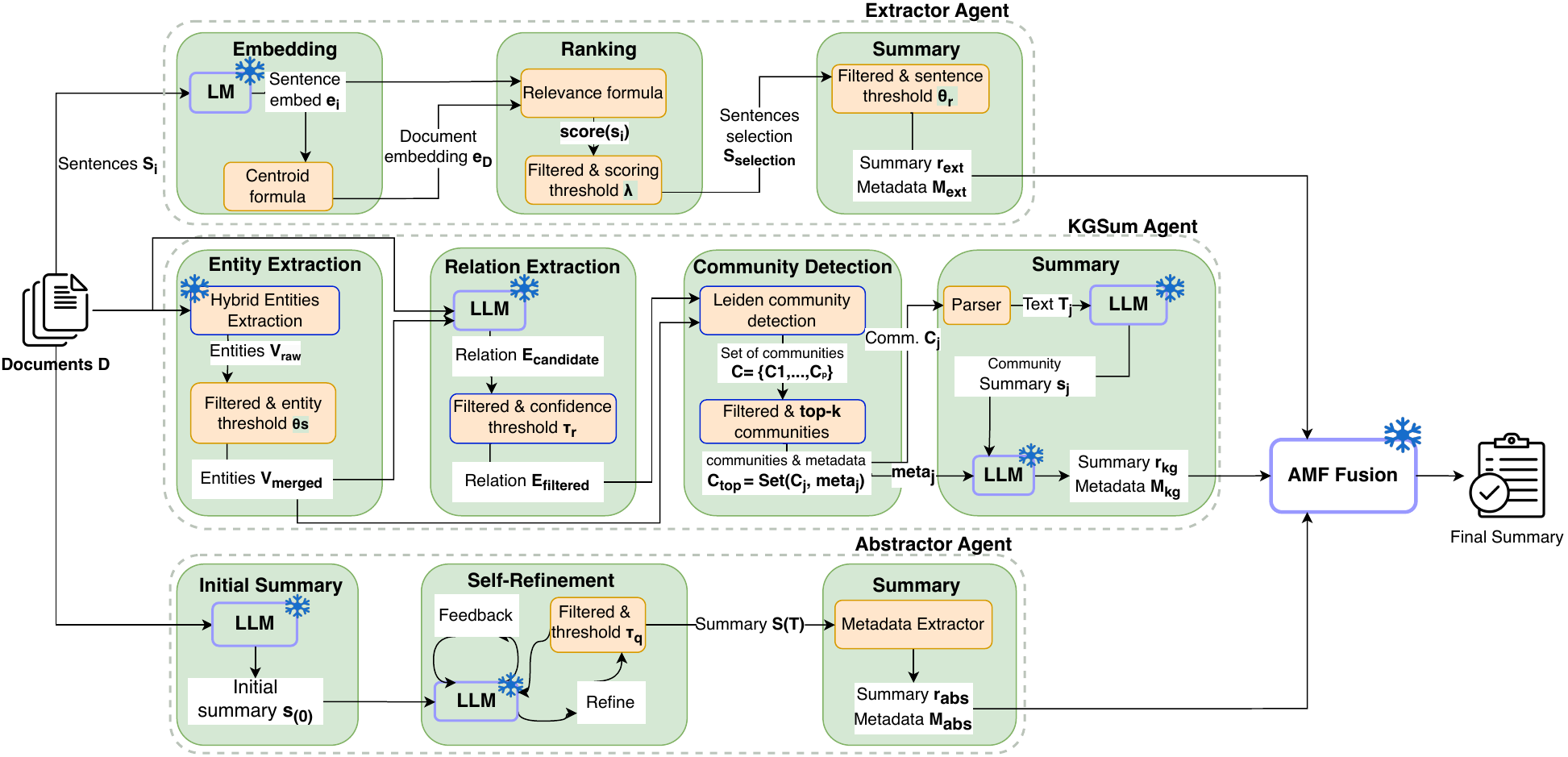}
  \caption{\textcolor{black}{Overall architecture of our proposed MoA framework. The three agents (Extractor, KGSum, Abstractor) process the input documents \(D\) through parallel specialized pipelines, generating their respective summaries (\(r_{\text{ext}}\), \(r_{\text{kg}}\), \(r_{\text{abs}}\)) and metadata (\(\mathcal{M}_{\text{ext}}\), \(\mathcal{M}_{\text{kg}}\), \(\mathcal{M}_{\text{abs}}\)). The AMF module then synthesizes these outputs into the final summary \(S_{\text{MoA}}\). ``LM'' denotes Language Model (BERT-based), while ``LLM'' denotes Large Language Model.}}
  \label{fig:overview-method}
\end{figure}

\section{Methodology}\label{sec:methodology}

\textcolor{black}{Figure~\ref{fig:overview-method} shows the overall framework architecture of MoA framework.}

\subsection{Extractor Agent}\label{subsec:Extractor-agent}

{\color{black}The Extractor Agent systematically identifies and extracts the most semantically significant sentences from the input document set \(D\) to generate an extractive summary \(r_{\text{ext}}\).} {\color{black}Building upon established centroid-based approaches \citep{rossiello_etal_2017_centroid}, our innovation integrates positional awareness with semantic centrality through a hybrid scoring mechanism that leverages pre-trained contextual embeddings.} {\color{black}As illustrated in the Extractor Agent section of Figure~\ref{fig:overview-method}, this process operates through three sequential stages: embedding generation, relevance scoring, and summary assembly.}

{\color{black}In the first stage, neural sentence embedding transforms each sentence \(s_i\) from document set \(D\) into contextual vector representations using a pre-trained Language Model (LM), specifically Longformer \citep{beltagy_longformer_2020}.} {\color{black}This encoding process produces sentence embeddings \(\mathbf{e}_i\) that capture semantic content within the document context.} {\color{black}Subsequently, the document-level semantic representation \(\mathbf{e}_D\) is computed as the centroid of all sentence embeddings:}
\begin{equation}\label{eq:document_embedding_extractor}
  {\color{black}\mathbf{e}_D = \frac{1}{m} \sum_{i=1}^m \mathbf{e}_i,}
\end{equation}
{\color{black}where \(m\) denotes the total number of sentences across all documents in \(D\).}

{\color{black}The second stage implements sentence ranking through a hybrid relevance scoring mechanism that balances semantic similarity with positional bias.} {\color{black}For each sentence \(s_i\), the relevance score integrates cosine similarity between sentence and document embeddings with positional preference:}
\begin{equation}\label{eq:sentence_score_extractor}
  {\color{black}\text{score}(s_i) = \lambda \cos(\mathbf{e}_i, \mathbf{e}_D) + (1-\lambda)\left(1 - \frac{\text{pos}(s_i)}{L_d}\right),}
\end{equation}
{\color{black}where \(\text{pos}(s_i)\) represents the normalized position of sentence \(s_i\) within the concatenated document set of length \(L_d\), and \(\lambda\) controls the semantic-positional balance, as shown in Table~\ref{tab:hyperparams}.} {\color{black}Sentences are subsequently ranked by these scores, with the top-\(\alpha\) percent forming the candidate set \(S_{\text{cand}}\).}

{\color{black}The third stage assembles the final extractive summary through redundancy filtering and metadata generation.} {\color{black}An optional redundancy removal step refines \(S_{\text{cand}}\) into \(S_{\text{sel}}\):}

\begin{equation}\label{eq:redundancy_filtering}
  {\color{black}S_{\text{sel}} = \{s_i \in S_{\text{cand}} : \forall s_j \in S_{\text{sel}}, \cos(\mathbf{e}_i, \mathbf{e}_j) \leq \theta_r\},}
\end{equation}
{\color{black}where \(\mathbf{e}_i\) and \(\mathbf{e}_j\) are the embeddings of sentences \(s_i\) and \(s_j\), respectively. And \(\theta_r\) denotes the redundancy threshold (Table~\ref{tab:hyperparams}).} {\color{black}Finally, the final summary \(r_{\text{ext}}\) concatenates selected sentences in their original document order, while the metadata package \(\mathcal{M}_{\text{ext}}\) encapsulates sentence scores, selection indices, and processing parameters for subsequent AMF integration.}

\subsection{KGSum Agent}\label{subsec:kgsum-agent}

\begin{figure}[htbp]
  \centering
	\includegraphics[width=0.6\linewidth]{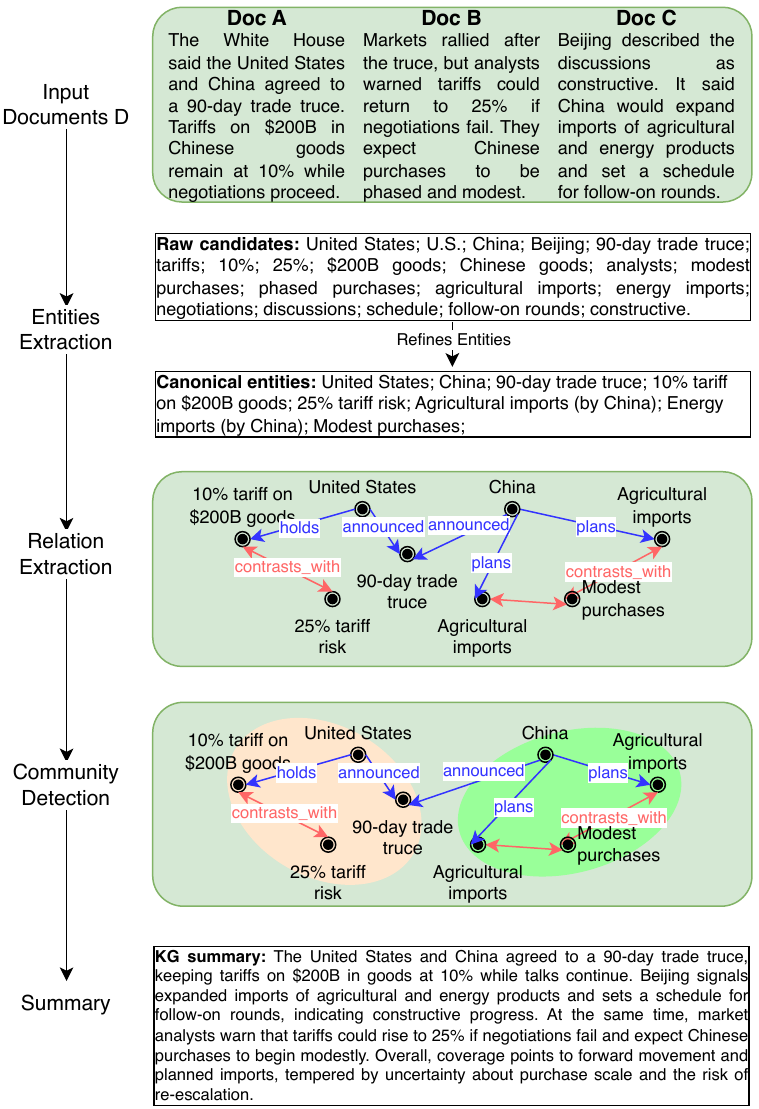}
  \caption{\textcolor{black}{Process of KGSum Agent for dynamic knowledge graph construction and knowledge-enhanced summary generation. Blue relation edges represent standard semantic relations, while red edges denote contrastive relations capturing conflicting information. The knowledge graph is dynamically constructed for each input document set, ensuring tailored representation of the specific content.}}
  \label{fig:kgsum-agent-example}
\end{figure}

{\color{black}The KGSum Agent dynamically constructs a knowledge graph from documents \(D\) to generate knowledge-enhanced summaries \(r_{\text{kg}}\).} {\color{black}Building upon established KG construction principles \citep{hogan2021knowledge}, our innovation extends traditional approaches with hybrid entity extraction, contrastive relation modeling, and meta-aware summarization, with detailed novelty analysis provided in Appendix~\ref{appendix:kgsum_detailed_analysis}.} {\color{black}The dynamic construction ensures that each new document set or revised data generates a fresh, tailored knowledge graph optimized for the specific input.} {\color{black}Figure~\ref{fig:overview-method} shows the overall architecture, while Figure~\ref{fig:kgsum-agent-example} demonstrates the processing pipeline.} {\color{black}The agent operates through four sequential stages, each building upon the previous stage's outputs.}

\subsubsection{Entity Extraction from Documents}\label{subsubsec:kg_entity_extraction}

{\color{black}From the document collection \(D\), the first stage employs hybrid entity extraction to identify key information units, as illustrated in the entity extraction phase of Figure~\ref{fig:kgsum-agent-example}.} {\color{black}We combine three complementary methods: spaCy NER for standard entities, rule-based patterns for numerical data, and LLM extraction with prompt \(\Pi_e\) (Figure~\ref{fig:entity_extraction_prompt}) for contextual understanding.} {\color{black}The raw entities are aggregated as:}
\begin{equation}\label{eq:entity_aggregation}
  {\color{black}V_{\text{raw}} = V_{\text{spacy}} \cup V_{\text{rules}} \cup V_{\text{llm}}.}
\end{equation}
{\color{black}Subsequently, entity consolidation eliminates redundancy through similarity-based merging, as demonstrated in the entity refinement step of Figure~\ref{fig:kgsum-agent-example}.} {\color{black}For entities \(v_i, v_j \in V_{\text{raw}}\), the merging process follows:}
\begin{equation}\label{eq:entity_merging}
  {\color{black}V_{\text{merged}} = \left\{
  \begin{array}{ll}
    ext{Merge}(v_i, v_j) & \text{if } \text{sim}(v_i, v_j) \geq \theta_s \\
    \{v_i, v_j\}           & \text{otherwise}
  \end{array}
  \right.,}
\end{equation}
{\color{black}where \(\theta_s\) represents the similarity threshold (Table~\ref{tab:hyperparams}).} {\color{black}This process yields the consolidated entity set \(V_{\text{merged}}\) for subsequent relation extraction.}

\subsubsection{Relation Extraction from Documents}\label{subsubsec:kg_relation_extraction}

{\color{black}Leveraging the consolidated entities \(V_{\text{merged}}\), the second stage identifies semantic connections to construct graph edges, as shown in the relation extraction phase of Figure~\ref{fig:kgsum-agent-example}.} {\color{black}Our approach focuses on extracting both standard semantic relations and contrastive relations that capture conflicting information across documents.} {\color{black}Using LLM with specialized prompt \(\Pi_{\text{rel}}\) (Figure~\ref{fig:relation_extraction_prompt}), the relation extraction process is formulated as:}
\begin{equation}\label{eq:relation_extraction}
  {\color{black}E_{\text{candidate}} = \text{LLM}(\Pi_{\text{rel}}, V_{\text{merged}}, D).}
\end{equation}
{\color{black}The prompt systematically guides extraction of traditional semantic relations alongside our novel contrastive relations, with empirical validation of contrastive relation effectiveness detailed in Figure~\ref{fig:relation_extraction_prompt}.} {\color{black}To ensure high-quality graph construction, confidence-based filtering is applied to each candidate relation \(r \in E_{\text{candidate}}\):}
\begin{equation}\label{eq:relation_filtering}
  {\color{black}E_{\text{filtered}} = \left\{
  \begin{array}{ll}
    r         & \text{if } \text{confidence}(r) \geq \tau_r \\
    \emptyset & \text{otherwise}
  \end{array}
  \right.,}
\end{equation}
{\color{black}where \(\tau_r\) represents the confidence threshold (Table~\ref{tab:hyperparams}).} {\color{black}This yields the refined relation set \(E_{\text{filtered}}\) that forms the knowledge graph edges.}

\subsubsection{Community Detection in Knowledge Graph}\label{subsubsec:kg_community_detection}

{\color{black}With the knowledge graph \(G = (V_{\text{merged}}, E_{\text{filtered}})\) constructed, the third stage partitions it into thematic communities for structured analysis, as illustrated in the community detection phase of Figure~\ref{fig:kgsum-agent-example}.} {\color{black}We employ the Leiden algorithm \citep{traag2019louvain} for community detection based on modularity optimization.} {\color{black}The partitioning process produces:}
\begin{equation}\label{eq:community_detection}
  {\color{black}\mathcal{C} = \{C_1, C_2, \ldots, C_p\} = \text{Leiden}(G).}
\end{equation}
{\color{black}Subsequently, metadata extraction generates descriptive information for each community \(C_j\):}
\begin{equation}\label{eq:metadata_extraction}
  {\color{black}\text{meta}_j = \{\text{centrality}, \text{contrast\_flags}, \text{key\_entities}, \text{fact\_density}\}.}
\end{equation}
{\color{black}This metadata encompasses centrality scores, contrastive relation indicators, key entities, and content density measures.} {\color{black}Finally, importance-based ranking selects the top-\(k\) communities (Table~\ref{tab:hyperparams}) to form \(\mathcal{C}_{\text{top}}\) for focused summarization.}

\subsubsection{KG-Based Summary Generation}\label{subsubsec:kg_summary_generation}

{\color{black}Building upon the selected communities \(\mathcal{C}_{\text{top}}\), the final stage generates the knowledge-enhanced summary \(r_{\text{kg}}\) through meta-aware fusion, as demonstrated in the summary generation phase of Figure~\ref{fig:kgsum-agent-example}.} {\color{black}Each community \(C_j \in \mathcal{C}_{\text{top}}\) undergoes linearization and targeted summarization:}
\begin{equation}\label{eq:community_summarization}
  {\color{black}T_j = \text{Linearize}(C_j), \quad s_j = \text{LLM}(\Pi_s, T_j),}
\end{equation}
{\color{black}where \(T_j\) converts the graph structure to text, and \(\Pi_s\) (Figure~\ref{fig:community_summary_prompt}) generates community-specific summaries.} {\color{black}Subsequently, meta-aware fusion integrates individual summaries while leveraging their associated metadata:}
\begin{equation}\label{eq:meta_aware_fusion}
  {\color{black}r_{\text{kg}} = \left\{\begin{array}{ll}
    	ext{LLM}(\Pi_f, \{(s_j, \text{meta}_j) : C_j \in \mathcal{C}_{\text{top}}\}) & \text{if } |\mathcal{C}_{\text{top}}| > 1 \\
    s_1                                                                            & \text{if } |\mathcal{C}_{\text{top}}| = 1
  \end{array}\right.,}
\end{equation}
{\color{black}The fusion prompt \(\Pi_f\) (Figure~\ref{fig:meta_aware_fusion_prompt}) utilizes metadata to weight communities by centrality, highlight conflicts, and ensure comprehensive coverage.} {\color{black}This process yields the final knowledge-enhanced summary \(r_{\text{kg}}\) along with metadata package \(\mathcal{M}_{\text{kg}}\) for subsequent AMF integration.}

\subsection{Abstractor Agent}\label{subsec:abstractor-agent}

{\color{black}The Abstractor Agent generates fluent, abstractive summaries by synthesizing information from documents \(D\) to produce \(r_{\text{abs}}\).} {\color{black}Building upon established LLM summarization techniques \citep{zhang2024benchmarking} and self-improvement concepts \citep{madaan2023self}, our innovation introduces an iterative self-refinement loop specifically designed for multi-document summarization.} {\color{black}As illustrated in the Abstractor Agent section of Figure~\ref{fig:overview-method}, this process operates through three sequential stages: initial summary generation, self-refinement, and final output production.}

{\color{black}In the first stage, initial summary generation transforms the document collection \(D\) into an abstractive summary using an LLM with specialized prompt \(\Pi_{\text{init}}\) (Figure~\ref{fig:abstractor_init_prompt}).} {\color{black}This process generates the initial summary:}
\begin{equation}\label{eq:initial_summary}
  {\color{black}\tilde{s}^{(0)} = \text{LLM}(\Pi_{\text{init}}, D),}
\end{equation}
{\color{black}where \(\tilde{s}^{(0)}\) captures the main themes from \(D\) and serves as the foundation for subsequent refinement.}

{\color{black}The second stage implements iterative self-refinement to enhance summary quality through feedback-driven improvement.} {\color{black}For each iteration \(t\), the system evaluates the current summary \(\tilde{s}^{(t)}\) using a multi-dimensional feedback function \(\mathcal{F}_{\text{MDS}}\) that assesses factual accuracy, coverage, coherence, and specificity.} {\color{black}The refinement process is formulated as:}
\begin{equation}\label{eq:self_refinement}
  {\color{black}\tilde{s}^{(t+1)} = \text{LLM}(\Pi_{\text{ref}}, \tilde{s}^{(t)}, \mathcal{F}_{\text{MDS}}(\tilde{s}^{(t)}, D)),}
\end{equation}
{\color{black}where \(\Pi_{\text{ref}}\) represents the refinement prompt (Figure~\ref{fig:abstractor_agent_prompt}) that guides quality improvement.} {\color{black}This iterative cycle continues for a maximum of \(T\) iterations or until the quality score exceeds threshold \(\tau_q\) (Table~\ref{tab:hyperparams}).}

{\color{black}The third stage produces the final abstractive summary and comprehensive metadata.} {\color{black}The refined summary becomes the final output \(r_{\text{abs}} = \tilde{s}^{(T)}\), while the metadata package \(\mathcal{M}_{\text{abs}}\) encapsulates quality scores, iteration count, and processing parameters.} {\color{black}Both \(r_{\text{abs}}\) and \(\mathcal{M}_{\text{abs}}\) are subsequently passed to the AMF module for multi-perspective integration.}

\subsection{Mixture of Agents Orchestration: The Adaptive Multi-Perspective Fusion (AMF) Mechanism}\label{subsec:moa_amf}

\begin{figure}[ht]
  \centering
	\includegraphics[width=1\textwidth]{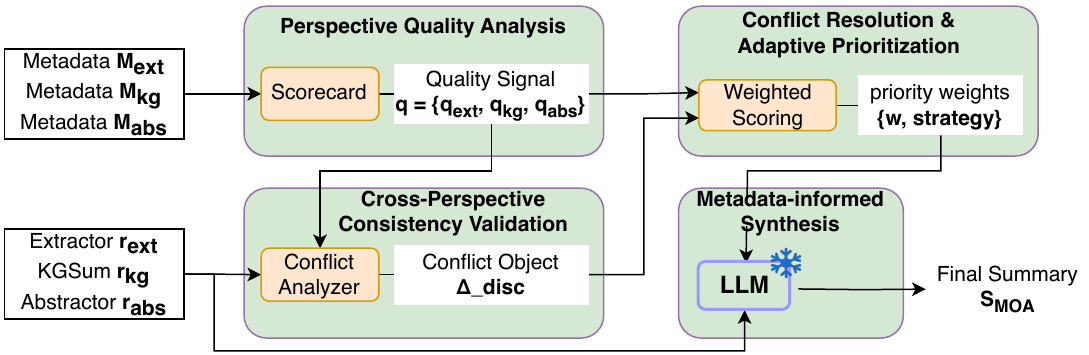}
  \caption{\textcolor{black}{The four-stage AMF pipeline for adaptive multi-perspective fusion. Stage 1 analyzes quality signals from each agent's metadata. Stage 2 validates cross-perspective consistency and detects conflicts. Stage 3 resolves conflicts and calculates adaptive priority weights. Stage 4 synthesizes the final summary using metadata-informed prompting. This architecture distinguishes our contribution from standard uniform-weighting MoA approaches.}}
  \label{fig:amf-module}
\end{figure}

{\color{black}The three agents produce specialized summaries that must be intelligently synthesized into the final output \(S_{\text{MoA}}\).} {\color{black}The Mixture of Agents paradigm \citep{wang2025mixtureofagents} combines multiple perspectives through iterative refinement.} {\color{black}However, existing MoA approaches apply uniform weighting to all agents, which ignores varying quality across inputs.} {\color{black}We propose Adaptive Multi-Perspective Fusion (AMF), which extends MoA with adaptive prioritization based on quality signals and domain strengths.} {\color{black}As shown in Figure~\ref{fig:overview-method}, AMF processes summaries \(r_{\text{ext}}\), \(r_{\text{kg}}\), \(r_{\text{abs}}\) and metadata \(\mathcal{M}_{\text{ext}}\), \(\mathcal{M}_{\text{kg}}\), \(\mathcal{M}_{\text{abs}}\) through four stages (Figure~\ref{fig:amf-module}).} {\color{black}This coordination mechanism distinguishes our architectural contribution from standard prompting approaches.}

\subsubsection{Stage 1: Perspective Quality Analysis}\label{subsubsec:amf_stage1}

{\color{black}Stage 1 extracts quality signals from agent metadata, as shown in Figure~\ref{fig:amf-module}.} {\color{black}Each agent response \(R_{\text{agent}} = \{r_{\text{agent}}, \mathcal{M}_{\text{agent}}\}\) already contains quality metrics computed during summary generation.} {\color{black}We filter relevant indicators from these metadata packages without additional computation.} {\color{black}For Extractor Agent, we extract from \(\mathcal{M}_{\text{ext}}\):}
\begin{equation}\label{eq:quality_extractor}
  {\color{black}\mathbf{q}_{\text{ext}} = \{q_{\text{salience}}, q_{\text{density}}, q_{\text{coverage}}\},}
\end{equation}
{\color{black}where \(q_{\text{salience}}\) reflects sentence importance, \(q_{\text{density}}\) measures factual content, and \(q_{\text{coverage}}\) assesses entity representation.}

{\color{black}Similarly, for KGSum Agent, we extract from \(\mathcal{M}_{\text{kg}}\):}
\begin{equation}\label{eq:quality_kg}
  {\color{black}\mathbf{q}_{\text{kg}} = \{q_{\text{relation}}, q_{\text{contrast}}, q_{\text{entity}}\},}
\end{equation}
{\color{black}where \(q_{\text{relation}}\) quantifies relation confidence, \(q_{\text{contrast}}\) captures contrastive flags, and \(q_{\text{entity}}\) measures coverage.}

{\color{black}For Abstractor Agent, we extract from \(\mathcal{M}_{\text{abs}}\):}
\begin{equation}\label{eq:quality_abs}
  {\color{black}\mathbf{q}_{\text{abs}} = \{q_{\text{fluency}}, q_{\text{coherence}}, q_{\text{synthesis}}\},}
\end{equation}
{\color{black}where \(q_{\text{fluency}}\) evaluates linguistic quality, \(q_{\text{coherence}}\) measures narrative flow, and \(q_{\text{synthesis}}\) assesses integration.} {\color{black}These filtered signals enable Stage 2 consistency validation.}

\subsubsection{Stage 2: Cross-Perspective Consistency Validation}\label{subsubsec:amf_stage2}

{\color{black}Building on the quality signals from Stage 1, Stage 2 validates consistency across summaries \(\{r_{\text{ext}}, r_{\text{kg}}, r_{\text{abs}}\}\), as shown in Figure~\ref{fig:amf-module}.} {\color{black}We extract numerical values and named entities from each summary using regex patterns and SpaCy NER.} {\color{black}For numerical consistency, we compute pairwise differences between extracted numbers across agents.} {\color{black}For entity consistency, we measure overlap using Jaccard similarity on entity sets.} {\color{black}The consistency analysis produces:}
\begin{equation}\label{eq:discrepancy}
  {\color{black}\Delta_{\text{disc}} = \{\text{contradictions}, \text{unique\_contributions}, \text{consistency\_score}\},}
\end{equation}
{\color{black}where \(\text{contradictions}\) lists conflicting facts between agents, \(\text{unique\_contributions}\) identifies perspective-specific insights, and \(\text{consistency\_score}\) aggregates numerical and entity agreement.} {\color{black}Specifically, consistency score is computed as:}
\begin{equation}\label{eq:consistency_score}
  {\color{black}\text{consistency\_score} = \frac{1}{3}\left(\text{num\_agreement} + \text{entity\_overlap} + \text{fact\_overlap}\right),}
\end{equation}
{\color{black}where each component ranges from 0 to 1.} {\color{black}This structured output guides Stage 3 prioritization.}

\subsubsection{Stage 3: Conflict Resolution and Adaptive Prioritization}\label{subsubsec:amf_stage3}

{\color{black}Stage 3 implements adaptive weighting, the core innovation distinguishing AMF from standard MoA (Figure~\ref{fig:amf-module}).} {\color{black}Using quality signals from Stage 1 and consistency analysis from Stage 2, we calculate agent-specific priority weights.} {\color{black}Unlike uniform weighting in existing MoA, our approach adjusts priorities dynamically based on input characteristics.} {\color{black}For each agent, the base quality score aggregates filtered signals:}
\begin{equation}\label{eq:enhanced_score}
  {\color{black}\text{score}_{\text{agent}} = \text{mean}(\mathbf{q}_{\text{agent}}),}
\end{equation}
{\color{black}where the mean aggregates quality signal components.}

{\color{black}Subsequently, we adjust scores with domain-specific bonuses and consistency-based penalties.} {\color{black}For KGSum Agent, we add bonuses when relation confidence exceeds thresholds.} {\color{black}For Extractor Agent, bonuses apply when salience scores are high.} {\color{black}For Abstractor Agent, bonuses reward fluency.} {\color{black}Moreover, agents with low consistency scores receive penalties.} {\color{black}The final priority weight is:}
\begin{equation}\label{eq:priority_weight}
  {\color{black}w_{\text{agent}} = \frac{\text{score}_{\text{agent}} + \text{bonus}_{\text{agent}} - \text{penalty}_{\text{agent}}}{\sum_{\text{all}} (\text{score} + \text{bonus} - \text{penalty})},}
\end{equation}
{\color{black}where normalization ensures \(\sum_{\text{agent}} w_{\text{agent}} = 1\).} {\color{black}These weights \(\mathbf{w} = \{w_{\text{ext}}, w_{\text{kg}}, w_{\text{abs}}\}\) prioritize high-quality agents for each input.}

{\color{black}Furthermore, we select fusion strategy based on weight distribution:}
\begin{equation}\label{eq:fusion_strategy}
  {\color{black}\text{strategy} = \begin{cases}
    	\text{KG\_LED}         & \text{if } w_{\text{kg}} > 0.5 \text{ and } q_{\text{relation}} > 0.7  \\
    	\text{EXTRACTOR\_LED}  & \text{if } w_{\text{ext}} > 0.5 \text{ and } q_{\text{salience}} > 0.7 \\
    	\text{ABSTRACTOR\_LED} & \text{if } w_{\text{abs}} > 0.5 \text{ and } q_{\text{fluency}} > 0.7  \\
    	\text{ADAPTIVE}        & \text{otherwise}
  \end{cases},}
\end{equation}
{\color{black}where thresholds determine when one agent dominates.} {\color{black}This strategy guides Stage 4 synthesis.}

\begin{figure}[ht]
  \centering
	\includegraphics[width=0.8\columnwidth]{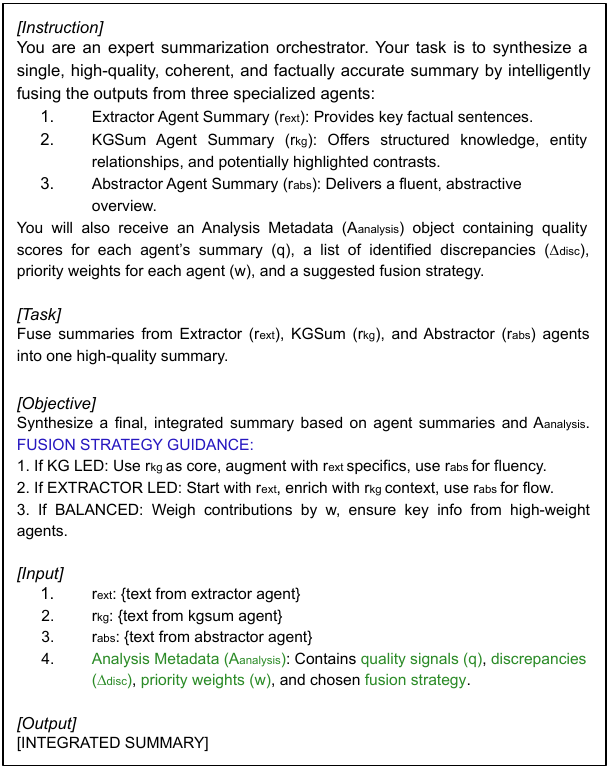}
  \caption{\textcolor{black}{Structure of the enhanced fusion prompt \(\Pi_{\text{AMF}}\) used in Stage 4 of AMF. The prompt dynamically incorporates quality signals \(\mathbf{q}\), priority weights \(\mathbf{w}\), discrepancy information \(\Delta_{\text{disc}}\), and the selected fusion strategy. This metadata-driven prompting enables the LLM to perform intelligent synthesis by explicitly prioritizing high-quality perspectives and resolving conflicts based on quantitative analysis.}}
  \label{fig:amf-fusion-prompt}
\end{figure}

\subsubsection{Stage 4: Metadata-Informed Synthesis}\label{subsubsec:amf_stage4}

{\color{black}Stage 4 synthesizes the final summary using metadata from Stages 1 to 3 (Figure~\ref{fig:amf-module}).} {\color{black}We consolidate all analysis components into:}
\begin{equation}\label{eq:analysis_metadata}
  {\color{black}\mathcal{A}_{\text{analysis}} = \{\mathbf{q}, \Delta_{\text{disc}}, \mathbf{w}, \text{strategy}\},}
\end{equation}
{\color{black}where quality signals, discrepancies, weights, and strategy guide synthesis.}

{\color{black}The fusion prompt \(\Pi_{\text{AMF}}\) embeds this metadata to direct LLM attention (Figure~\ref{fig:amf-fusion-prompt}).} {\color{black}Unlike standard prompting with equal attention, our prompt explicitly specifies priority weights and conflict resolution.} {\color{black}The final summary is:}
\begin{equation}\label{eq:final_synthesis}
  {\color{black}S_{\text{MoA}} = \text{LLM}(\Pi_{\text{AMF}}, \{r_{\text{ext}}, r_{\text{kg}}, r_{\text{abs}}\}, \mathcal{A}_{\text{analysis}}),}
\end{equation}
{\color{black}where the LLM receives summaries with explicit guidance from adaptive prioritization.} {\color{black}This architectural coordination mechanism demonstrates that performance gains stem from intelligent multi-agent orchestration rather than simple prompting.} {\color{black}AMF produces comprehensive metadata \(\mathcal{M}_{\text{AMF}}\) for transparency.}


\section{Experiments}\label{sec:experiments}

In this section, we present a comprehensive experimental evaluation of our proposed MoA framework for MDS. This evaluation is conducted on diverse real-world datasets. MoA is designed as a training-free, multi-agent architecture, while both SOTA full-training (supervised) and training-free (unsupervised) baselines are considered. Through rigorous comparisons with SOTA baselines and in-depth analyses, we aim to address the following research questions:
\begin{itemize}
    \item \textbf{RQ1:} How does the choice of LLM (open-source vs. closed-source) influence MoA's performance? Can MoA effectively leverage more advanced LLMs?
    \item \textbf{RQ2:} How does MoA perform on MDS tasks compared to SOTA supervised and unsupervised baselines?
    \item \textbf{RQ3:} What is the contribution of each agent within the MoA framework to the overall summarization quality?
\end{itemize}

We first investigate the impact of different LLM choices (both closed-source and open-source) on MoA's efficacy to answer RQ1. Based on these findings, we select the optimal LLM. We then establish MoA's performance against SOTA baselines (RQ2) over both English and Vietnamese datasets. Finally, an ablation study is carried out to dissect individual component contributions to answer RQ3.

\subsection{Experimental Setup}\label{subsec: setup-exp}

{\color{black}\textbf{Evaluation Datasets} We evaluate MoA on four diverse MDS datasets spanning multiple domains and languages: Multi-News (news articles, English) \citep{fabbri_multi-news_2019}, Multi-XScience (scientific papers, English) \citep{lu_multi-xscience_2020}, VN-MDS (news articles, Vietnamese)\footnote{https://github.com/lupanh/VietnameseMDS}, and ViMs (diverse documents, Vietnamese) \citep{tran2020vims}. Table~\ref{tab:dataset_stats} presents comprehensive statistics for all datasets, including test sample counts, number of reference summaries per cluster, average document lengths, and summary lengths. \textcolor{black}{Our evaluation encompasses substantial scale variation. Document quantities range from 29 samples (VN-MDS) to 5,622 samples (Multi-News), while average document lengths span from 419 words to 2,103 words. The number of reference summaries per cluster varies from 2.79 to 6.56, reflecting diverse multi-document aggregation requirements. Therefore, our evaluation validates architectural effectiveness across realistic scale variations encountered in established MDS benchmarks.} (Detailed descriptions of dataset characteristics, preprocessing procedures, and complexity analysis can be found in Appendix~\ref{appendix:dataset_details}.)}

\begin{table}[ht]
\centering
\caption{\textcolor{black}{Statistics of evaluation datasets}}
\label{tab:dataset_stats}
\begin{tabular*}{\columnwidth}{@{\extracolsep{\fill}}lccccc@{}}
\toprule
\textcolor{black}{\textbf{Dataset}} & \textcolor{black}{\textbf{Language}} & \textcolor{black}{\textbf{Test Samples}} & \textcolor{black}{\textbf{\# refs}} & \textcolor{black}{\textbf{doc. len}} & \textcolor{black}{\textbf{summ. len}} \\
\midrule
\textcolor{black}{Multi-News} & \textcolor{black}{English} & \textcolor{black}{5,622} & \textcolor{black}{2.79} & \textcolor{black}{2,103.49} & \textcolor{black}{263.66} \\
\textcolor{black}{Multi-XScience} & \textcolor{black}{English} & \textcolor{black}{5,093} & \textcolor{black}{4.42} & \textcolor{black}{778.08} & \textcolor{black}{116.44} \\
\textcolor{black}{VN-MDS} & \textcolor{black}{Vietnamese} & \textcolor{black}{29} & \textcolor{black}{3.14} & \textcolor{black}{419.0} & \textcolor{black}{172.1} \\
\textcolor{black}{ViMs} & \textcolor{black}{Vietnamese} & \textcolor{black}{45} & \textcolor{black}{6.56} & \textcolor{black}{470.2} & \textcolor{black}{231.2} \\
\bottomrule
\end{tabular*}
\end{table}

{\color{black}\textbf{Evaluation Metrics} Following previous works \citep{zhang_pegasus_2020}, we use ROUGE F1-scores as standard evaluation metrics: ROUGE-1 (R1) for content overlap, ROUGE-2 (R2) for fluency and local coherence, and ROUGE-L (RL) for structural similarity and overall coherence against reference summaries \citep{lin_rouge_2004}. Higher scores indicate better summarization quality.}

{\color{black}\textbf{Reporting Conventions} All ROUGE scores are reported as percentages. }{\color{red}In RQ2, MoA scores are reported as mean $\pm$ standard deviation over three runs, while baseline scores are taken from the original publications. RQ1 and RQ3 results are shown as mean scores only. Bold entries indicate the best score in a column, and \underline{underlined} entries indicate the second-best.} {\color{black}A hyphen ``-'' denotes that a method was not evaluated on that dataset or the original paper did not report that metric. Baseline scores are taken from the original publications. Abbreviations used in tables: FT = Full Training, TF = Training-Free, Prop. = Proprietary, Open = Open-source.}

{\color{black}\textbf{Implementation Details} }{\color{red}MoA is a training-free framework that can use different backbone LLMs, although its quality still depends on backbone suitability.} {\color{black}We systematically evaluate MoA with four leading closed-source LLMs (\texttt{GPT-4.1-mini}\footnote{https://openai.com/index/gpt-4-1/}, \texttt{Grok-3-mini}\footnote{https://x.ai/news/grok-3}, \texttt{GPT-4o-mini}\footnote{https://openai.com/index/hello-gpt-4o/}, \texttt{Gemini-1.5-flash-8b}\footnote{https://blog.google/technology/ai/google-gemini-next-generation-model-february-2024/}) and three representative open-source LLMs (\texttt{Llama-3.1-8B-Instruct}\footnote{https://huggingface.co/meta-llama/Llama-3.1-8B-Instruct}, \texttt{Qwen2.5-7B-Instruct}\footnote{https://huggingface.co/Qwen/Qwen2.5-7B-Instruct}, \texttt{Gemma-2-9b-it}\footnote{https://huggingface.co/google/gemma-2-9b-it}). All experiments are conducted on NVIDIA Quadro RTX 8000 GPU. We perform three independent runs for statistical validation, with mean and standard deviation reported. (Complete hyperparameter configurations and detailed parameter selection procedures are provided in Appendix~\ref{appendix:hyperparams}. API cost analysis for closed-source models is provided in Appendix~\ref{appendix:cost_analysis}.) For ablation studies, we utilize the few-shot prompt template $\Pi_{\text{few}}$ presented in Appendix~\ref{appendix:kgsum_prompts} for single LLM baseline comparisons.}

{\color{black}\textbf{Baseline Selection} We compare MoA against two types of baselines. Full-training (supervised) baselines include PEGASUS \citep{zhang_pegasus_2020}, LED \citep{beltagy_longformer_2020}, PRIMERA \citep{xiao_primera_2022}, and Centrum \citep{liu_centrum_2022} for English, and LatVisNewshead \citep{le2025latvis} for Vietnamese. Training-free baselines include SRI+beam \citep{xu_sri_2020} and GLIMMER-TTR \citep{estape_glimmer_2021} for English, and Graph combine Abstractive \citep{nguyen_graph_hybrid_2020} and Bert-VBD \citep{vuong2024bert} for Vietnamese. Training-free methods use pre-trained LLMs without additional fine-tuning on MDS datasets. Therefore, they do not require task-specific data collection or labeling, which is useful when specialized training data is limited or computational resources are constrained.}

\subsection{Impact of LLM Choice on MoA Performance (RQ1)}\label{subsec:llm_impact}

\textcolor{black}{\begin{table}[ht]
\centering
\caption{\textcolor{black}{Performance comparison of our proposed MoA with different LLM backbones on English datasets}}
\label{tab:llm_comparison_english}
\begin{tabular*}{\columnwidth}{@{\extracolsep{\fill}}l c ccc ccc@{}}
\toprule
& & \multicolumn{3}{c}{\textcolor{black}{\textbf{Multi-News}}} & \multicolumn{3}{c}{\textcolor{black}{\textbf{Multi-XScience}}} \\
\cmidrule(lr){3-5}\cmidrule(lr){6-8}
\textcolor{black}{\textbf{Method}} & \textcolor{black}{\textbf{Type}} & \textcolor{black}{\textbf{R1}} & \textcolor{black}{\textbf{R2}} & \textcolor{black}{\textbf{RL}} & \textcolor{black}{\textbf{R1}} & \textcolor{black}{\textbf{R2}} & \textcolor{black}{\textbf{RL}} \\
\midrule
\textcolor{black}{GPT-4.1-mini} & \textcolor{black}{Prop.} & \textcolor{black}{\textbf{54.97}} & \textcolor{black}{\underline{24.90}} & \textcolor{black}{\textbf{29.40}} & \textcolor{black}{\textbf{36.66}} & \textcolor{black}{\textbf{5.64}} & \textcolor{black}{\textbf{19.73}} \\
\textcolor{black}{Grok-3-mini} & \textcolor{black}{Prop.} & \textcolor{black}{\underline{53.91}} & \textcolor{black}{24.68} & \textcolor{black}{28.61} & \textcolor{black}{\underline{35.60}} & \textcolor{black}{\underline{5.42}} & \textcolor{black}{\underline{18.94}} \\
\textcolor{black}{GPT-4o-mini} & \textcolor{black}{Prop.} & \textcolor{black}{52.67} & \textcolor{black}{23.72} & \textcolor{black}{27.97} & \textcolor{black}{31.05} & \textcolor{black}{4.72} & \textcolor{black}{16.28} \\
\textcolor{black}{Gemini-1.5-flash-8b} & \textcolor{black}{Prop.} & \textcolor{black}{51.92} & \textcolor{black}{23.10} & \textcolor{black}{27.51} & \textcolor{black}{30.11} & \textcolor{black}{4.50} & \textcolor{black}{15.82} \\
\textcolor{black}{Llama-3.1-8b-Instruct} & \textcolor{black}{Open} & \textcolor{black}{52.40} & \textcolor{black}{\textbf{25.98}} & \textcolor{black}{\underline{28.90}} & \textcolor{black}{23.42} & \textcolor{black}{2.60} & \textcolor{black}{13.24} \\
\textcolor{black}{Qwen2.5-7B-Instruct} & \textcolor{black}{Open} & \textcolor{black}{48.17} & \textcolor{black}{21.75} & \textcolor{black}{26.55} & \textcolor{black}{23.64} & \textcolor{black}{2.55} & \textcolor{black}{12.63} \\
\textcolor{black}{Gemma-2-9b-it} & \textcolor{black}{Open} & \textcolor{black}{25.08} & \textcolor{black}{8.48} & \textcolor{black}{14.81} & \textcolor{black}{24.05} & \textcolor{black}{2.47} & \textcolor{black}{12.84} \\
\bottomrule
\end{tabular*}%
\end{table}}

\textcolor{black}{\begin{table}[ht]
\centering
\caption{\textcolor{black}{Performance comparison of our proposed MoA with different LLM backbones on Vietnamese datasets}}
\label{tab:llm_comparison_vietnamese}
\begin{tabular*}{\columnwidth}{@{\extracolsep{\fill}}l c ccc ccc@{}}
\toprule
& & \multicolumn{3}{c}{\textcolor{black}{\textbf{VN-MDS}}} & \multicolumn{3}{c}{\textcolor{black}{\textbf{ViMs}}} \\
\cmidrule(lr){3-5}\cmidrule(lr){6-8}
\textcolor{black}{\textbf{Method}} & \textcolor{black}{\textbf{Type}} & \textcolor{black}{\textbf{R1}} & \textcolor{black}{\textbf{R2}} & \textcolor{black}{\textbf{RL}} & \textcolor{black}{\textbf{R1}} & \textcolor{black}{\textbf{R2}} & \textcolor{black}{\textbf{RL}} \\
\midrule
\textcolor{black}{GPT-4.1-mini} & \textcolor{black}{Prop.} & \textcolor{black}{\textbf{80.11}} & \textcolor{black}{\textbf{52.32}} & \textcolor{black}{\textbf{48.29}} & \textcolor{black}{\textbf{80.26}} & \textcolor{black}{\textbf{55.57}} & \textcolor{black}{\textbf{49.04}} \\
\textcolor{black}{Grok-3-mini} & \textcolor{black}{Prop.} & \textcolor{black}{\underline{79.05}} & \textcolor{black}{\underline{52.10}} & \textcolor{black}{\underline{47.50}} & \textcolor{black}{\underline{79.20}} & \textcolor{black}{\underline{55.35}} & \textcolor{black}{\underline{48.25}} \\
\textcolor{black}{GPT-4o-mini} & \textcolor{black}{Prop.} & \textcolor{black}{72.90} & \textcolor{black}{46.10} & \textcolor{black}{40.50} & \textcolor{black}{68.80} & \textcolor{black}{49.50} & \textcolor{black}{39.00} \\
\textcolor{black}{Gemini-1.5-flash-8b} & \textcolor{black}{Prop.} & \textcolor{black}{72.25} & \textcolor{black}{45.60} & \textcolor{black}{40.02} & \textcolor{black}{68.05} & \textcolor{black}{48.80} & \textcolor{black}{38.42} \\
\textcolor{black}{Llama-3.1-8b-Instruct} & \textcolor{black}{Open} & \textcolor{black}{71.80} & \textcolor{black}{45.20} & \textcolor{black}{39.80} & \textcolor{black}{67.50} & \textcolor{black}{48.50} & \textcolor{black}{38.50} \\
\textcolor{black}{Qwen2.5-7B-Instruct} & \textcolor{black}{Open} & \textcolor{black}{65.10} & \textcolor{black}{40.50} & \textcolor{black}{38.80} & \textcolor{black}{61.00} & \textcolor{black}{41.00} & \textcolor{black}{34.00} \\
\textcolor{black}{Gemma-2-9b-it} & \textcolor{black}{Open} & \textcolor{black}{63.20} & \textcolor{black}{38.70} & \textcolor{black}{37.10} & \textcolor{black}{59.00} & \textcolor{black}{39.00} & \textcolor{black}{32.00} \\
\bottomrule
\end{tabular*}%
\end{table}}

\textcolor{black}{Tables \ref{tab:llm_comparison_english} and \ref{tab:llm_comparison_vietnamese} present comprehensive performance comparison of MoA with different LLM backbones across all four evaluation datasets. Since MoA operates as a multi-agent system where LLMs serve as the primary reasoning engines for each agent, we systematically investigate how different LLM capabilities affect overall architectural performance. The results reveal clear performance hierarchies and important insights about LLM selection for MDS tasks.}

\textcolor{black}{On English datasets (Table~\ref{tab:llm_comparison_english}), closed-source models demonstrate substantial superiority over open-source alternatives. Specifically, \texttt{GPT-4.1-mini} achieves the highest scores across all metrics: 54.97 R1, 24.90 R2, and 29.40 RL on Multi-News, along with 36.66 R1, 5.64 R2, and 19.73 RL on Multi-XScience. These scores surpass \texttt{Grok-3-mini} by 2.0 percent in R1 on Multi-News and 3.0 percent on Multi-XScience, demonstrating consistent leadership. Moreover, \texttt{GPT-4o-mini} and \texttt{Gemini-1.5-flash-8b} achieve competitive results with 52.67 R1 and 51.92 R1 on Multi-News respectively, validating their strong general capabilities. In particular, among open-source LLMs, \texttt{Llama-3.1-8b-Instruct} achieves the best performance with 52.40 R1 on Multi-News, substantially outperforming other open-source alternatives. However, \texttt{Gemma-2-9b-it}, designed as a multimodal model, shows suboptimal performance with only 25.08 R1 on Multi-News, indicating that specialized capabilities do not always transfer effectively to text-only MDS applications.}

\textcolor{black}{On Vietnamese datasets (Table~\ref{tab:llm_comparison_vietnamese}), the performance hierarchy remains consistent with English results. \texttt{GPT-4.1-mini} achieves 80.11 R1 on VN-MDS and 80.26 R1 on ViMs, closely followed by \texttt{Grok-3-mini} with 79.05 R1 and 79.20 R1 respectively. The minimal performance gap (1.3 percent on VN-MDS, 1.3 percent on ViMs) confirms both models' strong multilingual capabilities. Furthermore, the performance differences between Vietnamese and English datasets primarily reflect dataset complexity characteristics rather than language-specific advantages. Specifically, Vietnamese datasets exhibit significantly higher extractiveness (94.3 percent versus 19.2 percent for English) and content redundancy, inflating ROUGE scores through exact n-gram matching rather than demanding abstractive synthesis. Detailed comparative complexity analysis with supporting metrics is presented in Table~\ref{tab:dataset_complexity_comparison} of Appendix~\ref{appendix:dataset_complexity}.}

\textcolor{red}{The results show that MoA performs best with strong closed-source models, while suitable instruction-tuned open-source models such as \texttt{Llama-3.1-8B-Instruct} and \texttt{Qwen2.5-7B-Instruct} still achieve competitive performance. By contrast, when MoA uses a less suitable backbone such as \texttt{Gemma-2-9b-it}, the overall summarization quality decreases noticeably. Therefore, \texttt{GPT-4.1-mini} is selected for the main comparative evaluation in RQ2 and the controlled ablation study in RQ3.}

\subsection{Comparative Performance Evaluation (RQ2)}\label{subsec:comparative_performance}

\textcolor{black}{To address RQ2, we evaluate MoA's performance against established full training and training-free baselines. For these comparisons, MoA utilizes \texttt{GPT-4.1-mini} as its backbone LLM based on its superior overall performance demonstrated in RQ1 (Section \ref{subsec:llm_impact}). Furthermore, we conduct three independent experimental runs to ensure result reliability, with mean and standard deviation reported for our method. Moreover, due to computational resource constraints and API cost considerations for large-scale evaluation across multiple datasets, we perform three runs as a practical balance between statistical validation and feasibility. Comprehensive results are presented in Tables \ref{tab:english_summary_results} and \ref{tab:vietnamese_summary_results} for English and Vietnamese datasets, respectively.}

\textcolor{black}{\begin{table}[ht]
\centering
\caption{\textcolor{black}{Performance comparison on English summarization datasets}}
\label{tab:english_summary_results}
\begin{tabular*}{\columnwidth}{@{\extracolsep{\fill}}l c ccc ccc}
\toprule
& & \multicolumn{3}{c}{\textcolor{black}{\textbf{Multi-News}}} & \multicolumn{3}{c}{\textcolor{black}{\textbf{Multi-XScience}}} \\
\cmidrule(lr){3-5}\cmidrule(lr){6-8}
\textcolor{black}{\textbf{Method}} & \textcolor{black}{\textbf{Type}} & \textcolor{black}{\textbf{R1}} & \textcolor{black}{\textbf{R2}} & \textcolor{black}{\textbf{RL}} & \textcolor{black}{\textbf{R1}} & \textcolor{black}{\textbf{R2}} & \textcolor{black}{\textbf{RL}} \\
\midrule
\textcolor{black}{PEGASUS} & \textcolor{black}{FT} & \textcolor{black}{32.0} & \textcolor{black}{10.0} & \textcolor{black}{17.0} & \textcolor{black}{28.0} & \textcolor{black}{\underline{5.0}} & \textcolor{black}{15.0} \\
\textcolor{black}{LED} & \textcolor{black}{FT} & \textcolor{black}{17.0} & \textcolor{black}{4.0} & \textcolor{black}{10.0} & \textcolor{black}{15.0} & \textcolor{black}{2.0} & \textcolor{black}{10.0} \\
\textcolor{black}{PRIMERA} & \textcolor{black}{FT} & \textcolor{black}{42.0} & \textcolor{black}{14.0} & \textcolor{black}{21.0} & \textcolor{black}{29.0} & \textcolor{black}{\underline{5.0}} & \textcolor{black}{16.0} \\
\textcolor{black}{Centrum} & \textcolor{black}{FT} & \textcolor{black}{\underline{46.0}} & \textcolor{black}{\underline{17.0}} & \textcolor{black}{\underline{23.0}} & \textcolor{black}{-} & \textcolor{black}{-} & \textcolor{black}{-} \\
\textcolor{black}{SRI+beam} & \textcolor{black}{TF} & \textcolor{black}{44.0} & \textcolor{black}{15.0} & \textcolor{black}{19.0} & \textcolor{black}{-} & \textcolor{black}{-} & \textcolor{black}{-} \\
\textcolor{black}{GLIMMER-TTR} & \textcolor{black}{TF} & \textcolor{black}{44.0} & \textcolor{black}{15.0} & \textcolor{black}{21.0} & \textcolor{black}{\underline{32.0}} & \textcolor{black}{\underline{5.0}} & \textcolor{black}{\underline{17.0}} \\
\textcolor{black}{\textbf{MoA (Ours)}} & \textcolor{black}{TF} & \textcolor{black}{\textbf{55.0$\pm$0.3}} & \textcolor{black}{\textbf{24.9$\pm$0.4}} & \textcolor{black}{\textbf{29.4$\pm$0.3}} & \textcolor{black}{\textbf{36.7$\pm$0.2}} & \textcolor{black}{\textbf{5.6$\pm$0.1}} & \textcolor{black}{\textbf{19.7$\pm$0.2}} \\
\bottomrule
\end{tabular*}
\end{table}}

\textcolor{black}{\begin{table}[ht]
\centering
\caption{\textcolor{black}{Performance comparison on Vietnamese summarization datasets}}
\label{tab:vietnamese_summary_results}
\begin{tabular*}{\columnwidth}{@{\extracolsep{\fill}}l c ccc ccc}
\toprule
& & \multicolumn{3}{c}{\textcolor{black}{\textbf{VN-MDS}}} & \multicolumn{3}{c}{\textcolor{black}{\textbf{ViMs}}} \\
\cmidrule(lr){3-5}\cmidrule(lr){6-8}
\textcolor{black}{\textbf{Method}} & \textcolor{black}{\textbf{Type}} & \textcolor{black}{\textbf{R1}} & \textcolor{black}{\textbf{R2}} & \textcolor{black}{\textbf{RL}} & \textcolor{black}{\textbf{R1}} & \textcolor{black}{\textbf{R2}} & \textcolor{black}{\textbf{RL}} \\
\midrule
\textcolor{black}{Graph combine Abs} & \textcolor{black}{TF} & \textcolor{black}{69.0} & \textcolor{black}{35.0} & \textcolor{black}{33.0} & \textcolor{black}{-} & \textcolor{black}{-} & \textcolor{black}{-} \\
\textcolor{black}{Bert-VBD} & \textcolor{black}{TF} & \textcolor{black}{70.0} & \textcolor{black}{40.0} & \textcolor{black}{34.0} & \textcolor{black}{-} & \textcolor{black}{-} & \textcolor{black}{-} \\
\textcolor{black}{LatVisNewshead} & \textcolor{black}{FT} & \textcolor{black}{\underline{77.0}} & \textcolor{black}{\underline{50.0}} & \textcolor{black}{\underline{47.0}} & \textcolor{black}{\underline{79.0}} & \textcolor{black}{\underline{55.0}} & \textcolor{black}{\underline{48.0}} \\
\textcolor{black}{\textbf{MoA (Ours)}} & \textcolor{black}{TF} & \textcolor{black}{\textbf{80.1}}$\pm$\textcolor{black}{\textbf{0.5}} & \textcolor{black}{\textbf{52.3}}$\pm$\textcolor{black}{\textbf{0.7}} & \textcolor{black}{\textbf{48.3}}$\pm$\textcolor{black}{\textbf{0.6}} & \textcolor{black}{\textbf{80.3}}$\pm$\textcolor{black}{\textbf{0.5}} & \textcolor{black}{\textbf{55.6}}$\pm$\textcolor{black}{\textbf{0.8}} & \textcolor{black}{\textbf{49.0}}$\pm$\textcolor{black}{\textbf{0.6}} \\
\bottomrule
\end{tabular*}%
\end{table}}

\textcolor{black}{Tables \ref{tab:english_summary_results} and \ref{tab:vietnamese_summary_results} present comprehensive performance comparison across all four evaluation datasets. The results demonstrate MoA's consistent superiority over existing methods across diverse evaluation settings.}

\textcolor{black}{On English datasets, our framework establishes new state-of-the-art results as shown in Table~\ref{tab:english_summary_results}. On Multi-News, MoA achieves 55.0$\pm$0.3 R1, 24.9$\pm$0.4 R2, and 29.4$\pm$0.3 RL. These scores substantially outperform both supervised and training-free baselines. Specifically, MoA improves over Centrum by 19.6 percent (from 46.0 to 55.0) in R1 and surpasses GLIMMER-TTR by 25.0 percent in R1. On Multi-XScience, MoA achieves 36.7$\pm$0.2 R1, surpassing GLIMMER-TTR by 14.7 percent. Moreover, MoA demonstrates substantial improvements in R2 and RL metrics: 46.5 percent improvement over Centrum in R2 (from 17.0 to 24.9) and 27.8 percent in RL (from 23.0 to 29.4) on Multi-News. This validates robust generalization to specialized scientific domains where precise terminology and complex relationships require sophisticated coordination.}

\textcolor{black}{The framework demonstrates language-agnostic effectiveness on Vietnamese datasets as presented in Table~\ref{tab:vietnamese_summary_results}. On VN-MDS, MoA achieves 80.1$\pm$0.5 R1, 52.3$\pm$0.7 R2, and 48.3$\pm$0.6 RL. On ViMs, MoA reaches 80.3$\pm$0.5 R1, 55.6$\pm$0.8 R2, and 49.0$\pm$0.6 RL. In particular, MoA surpasses the specialized Vietnamese model LatVisNewshead by 4.0 percent in R1 on VN-MDS (from 77.0 to 80.1) and 1.6 percent on ViMs (from 79.0 to 80.3) despite operating in training-free mode. Furthermore, MoA achieves consistent improvements across all metrics: 4.6 percent in R2 and 2.8 percent in RL on VN-MDS, and 1.1 percent in R2 and 2.1 percent in RL on ViMs. This validates that our architectural coordination approach successfully leverages multilingual LLM capabilities without language-specific fine-tuning.} \textcolor{red}{Tables~\ref{tab:llm_comparison_english} and~\ref{tab:llm_comparison_vietnamese} further show that MoA remains competitive with several reported baselines when it uses suitable open-source backbones such as \texttt{Llama-3.1-8B-Instruct} and \texttt{Qwen2.5-7B-Instruct}. In contrast, the results of \texttt{Gemma-2-9b-it} show that MoA becomes less competitive when the backbone is less suitable for this task.}

\textcolor{black}{Consistent improvements across diverse document types further validate architectural robustness. The framework handles news articles (Multi-News, VN-MDS), scientific papers (Multi-XScience), and diverse documents (ViMs) with systematic performance gains ranging from 1.6 percent to 46.5 percent across different metrics and datasets. Moreover, statistical validation presented in Appendix~\ref{appendix:statistical_validation} confirms that MoA significantly outperforms all baselines across 12 dataset-metric combinations with $p < 0.001$. Therefore, the multi-agent architecture with knowledge graph reasoning and adaptive fusion provides generalizable solutions for complex inter-document relationship modeling.}

\subsection{Ablation Study (RQ3)}\label{subsec:rq3_ablation}

\textcolor{black}{\begin{table}[ht]
\caption{\textcolor{black}{Ablation study results for our proposed MoA using GPT-4.1-mini}}
\label{tab:ablation_study_results}
\centering
\begin{tabular*}{\textwidth}{@{\extracolsep{\fill}}lcccccc}
\toprule
& \multicolumn{3}{c}{\textcolor{black}{\textbf{Multi-News (English)}}} & \multicolumn{3}{c}{\textcolor{black}{\textbf{VN-MDS (Vietnamese)}}} \\
\cmidrule(lr){2-4}\cmidrule(lr){5-7}
\textcolor{black}{\textbf{Configuration}} & \textcolor{black}{\textbf{R1}} & \textcolor{black}{\textbf{R2}} & \textcolor{black}{\textbf{RL}} & \textcolor{black}{\textbf{R1}} & \textcolor{black}{\textbf{R2}} & \textcolor{black}{\textbf{RL}} \\
\midrule
\multicolumn{7}{l}{\textcolor{black}{\textit{Part 1: Single-Agent Performance and LLM Baseline}}} \\
\textcolor{black}{LLM with structured prompt} & \textcolor{black}{27.0} & \textcolor{black}{5.0} & \textcolor{black}{11.0} & \textcolor{black}{39.3} & \textcolor{black}{6.6} & \textcolor{black}{15.2} \\
\textcolor{black}{Abstractor agent only} & \textcolor{black}{31.0} & \textcolor{black}{3.0} & \textcolor{black}{12.0} & \textcolor{black}{45.2} & \textcolor{black}{3.9} & \textcolor{black}{16.6} \\
\textcolor{black}{Extractor agent only} & \textcolor{black}{33.0} & \textcolor{black}{6.0} & \textcolor{black}{17.0} & \textcolor{black}{48.1} & \textcolor{black}{7.9} & \textcolor{black}{23.5} \\
\textcolor{black}{KGSum agent only} & \textcolor{black}{37.0} & \textcolor{black}{4.0} & \textcolor{black}{12.0} & \textcolor{black}{59.8} & \textcolor{black}{9.2} & \textcolor{black}{26.2} \\
\midrule
\multicolumn{7}{l}{\textcolor{black}{\textit{Part 2: Multi-Agent Combinations and Full Framework}}} \\
\textcolor{black}{MoA w/o KGSum agent} & \textcolor{black}{48.9} & \textcolor{black}{21.1} & \textcolor{black}{25.9} & \textcolor{black}{70.0} & \textcolor{black}{44.7} & \textcolor{black}{44.3} \\
\textcolor{black}{MoA w/o Abstractor agent} & \textcolor{black}{\underline{51.5}} & \textcolor{black}{\underline{23.2}} & \textcolor{black}{\underline{27.3}} & \textcolor{black}{\underline{73.8}} & \textcolor{black}{\underline{48.2}} & \textcolor{black}{\underline{45.8}} \\
\textcolor{black}{MoA w/o Extractor agent} & \textcolor{black}{44.8} & \textcolor{black}{18.7} & \textcolor{black}{23.3} & \textcolor{black}{65.1} & \textcolor{black}{38.1} & \textcolor{black}{41.9} \\
\textcolor{black}{\textbf{Full MoA framework}} & \textcolor{black}{\textbf{55.0}} & \textcolor{black}{\textbf{24.9}} & \textcolor{black}{\textbf{29.4}} & \textcolor{black}{\textbf{80.1}} & \textcolor{black}{\textbf{52.3}} & \textcolor{black}{\textbf{48.3}} \\
\bottomrule
\end{tabular*}
\end{table}}

\textcolor{black}{Table~\ref{tab:ablation_study_results} presents comprehensive ablation study results investigating the individual contribution of each agent within the MoA framework to address RQ3. We systematically evaluate single-agent performance, multi-agent combinations, and the full framework using \texttt{GPT-4.1-mini} on Multi-News and VN-MDS datasets.} \textcolor{red}{Here, the backbone LLM is intentionally fixed so that RQ3 can focus on the contribution of KGSum, Abstractor, and Extractor within one consistent MoA setting.}

\textcolor{black}{KGSum agent demonstrates strongest single-agent performance. Among individual agents, KGSum achieves 37.0 R1 on Multi-News and 59.8 R1 on VN-MDS, substantially outperforming Extractor (33.0 R1, 48.1 R1) and Abstractor (31.0 R1, 45.2 R1). Moreover, KGSum surpasses the simple LLM baseline (27.0 R1, 39.3 R1) by 37.0 percent and 52.2 percent respectively. This validates the effectiveness of knowledge graph construction and community-based reasoning for capturing inter-document relationships, confirming KGSum as the core architectural component for MDS tasks.}

\textcolor{black}{Multi-agent combinations consistently outperform single agents. All two-agent configurations achieve superior performance compared to individual agents, demonstrating complementary agent interactions. Specifically, MoA without Abstractor reaches 51.5 R1 on Multi-News, while MoA without KGSum achieves 48.9 R1. In particular, removing KGSum produces the worst degradation: MoA without KGSum scores only 48.9 R1 versus 51.5 R1 for MoA without Abstractor, an 11.2 percent performance gap. Furthermore, MoA without Extractor achieves 44.8 R1, demonstrating that factual content extraction remains essential for coordination. These patterns validate that KGSum provides critical structural understanding that other agents cannot compensate for, emphasizing its central role in multi-agent coordination.}

\textcolor{black}{Full MoA framework with AMF achieves optimal performance. The complete three-agent MoA reaches 55.0 R1 on Multi-News and 80.1 R1 on VN-MDS, outperforming all ablated configurations. The full framework improves over the best two-agent combination (MoA without Abstractor: 51.5 R1, 73.8 R1) by 6.8 percent and 8.5 percent respectively. Therefore, the AMF mechanism effectively integrates complementary perspectives from all three agents, while the complete MoA architecture provides a generalizable solution for complex multi-document summarization through coordinated knowledge graph reasoning and adaptive multi-agent fusion.}

\subsection{Human Evaluation}\label{subsec:human_evaluation}

\textcolor{black}{To verify the practical quality of summaries generated by our MoA framework, we conduct human evaluation beyond automatic metrics. Automatic metrics like ROUGE measure lexical overlap but cannot fully capture summary fluency and coherence. Therefore, systematic human judgment provides essential validation of real-world summary quality.}

\textcolor{black}{Following established evaluation guidelines from DUC 2007\footnote{https://www-nlpir.nist.gov/projects/duc/duc2007/quality-questions.txt}, we assess summaries across four key dimensions: grammaticality (linguistic correctness and sentence structure quality), non-redundancy (absence of repetitive information), referential clarity (pronoun and entity reference understandability), and structure coherence (logical organization and information flow). These four dimensions provide comprehensive coverage of summary quality from human perspective.}

\textcolor{black}{For experimental settings, we randomly select 20 document clusters from Multi-News for English and 20 clusters from VN-MDS for Vietnamese. We recruit 100 participants to evaluate generated summaries. Each participant rates summaries on a scale from 1 to 5 for each dimension, where higher scores indicate better quality. The annotation interface and detailed evaluation criteria are presented in Appendix~\ref{appendix:human_eval_interface}. This setup balances evaluation coverage with practical feasibility for large-scale assessment. Table~\ref{tab:human_eval_english} presents human evaluation results on Multi-News comparing MoA against two strong baselines (Centrum and PRIMERA), while Table~\ref{tab:human_eval_vietnamese} shows results on VN-MDS comparing MoA against BertVBD.}

\textcolor{black}{\begin{table}[ht]
\centering
\caption{\textcolor{black}{Human evaluation results on Multi-News dataset (English)}}
\label{tab:human_eval_english}
\begin{tabular*}{\columnwidth}{@{\extracolsep{\fill}}lcccc@{}}
\toprule
\textcolor{black}{\textbf{Model}} & \textcolor{black}{\textbf{Gram.}} & \textcolor{black}{\textbf{Non-red.}} & \textcolor{black}{\textbf{Ref. Clar.}} & \textcolor{black}{\textbf{Str. \& Coh.}} \\
\midrule
\textcolor{black}{Centrum} & \textcolor{black}{3.57} & \textcolor{black}{3.35} & \textcolor{black}{3.67} & \textcolor{black}{3.37} \\
\textcolor{black}{PRIMERA} & \textcolor{black}{2.75} & \textcolor{black}{3.15} & \textcolor{black}{3.10} & \textcolor{black}{2.75} \\
\textcolor{black}{\textbf{MoA (Ours)}} & \textcolor{black}{\textbf{4.56}} & \textcolor{black}{\textbf{4.67}} & \textcolor{black}{\textbf{4.55}} & \textcolor{black}{\textbf{4.47}} \\
\bottomrule
\end{tabular*}
\end{table}}

\textcolor{black}{Table~\ref{tab:human_eval_english} presents human evaluation results on Multi-News dataset. We compare MoA against two strong baselines: Centrum (supervised full-training) and PRIMERA (supervised). MoA achieves consistent superiority across all four evaluation dimensions.}

\textcolor{black}{MoA demonstrates strong performance across all quality dimensions. Specifically, MoA achieves 4.56 in grammaticality, 4.67 in non-redundancy, 4.55 in referential clarity, and 4.47 in structure coherence. These scores substantially outperform both baselines. In particular, MoA improves non-redundancy by 39.4 percent over Centrum (from 3.35 to 4.67). Moreover, MoA achieves 27.7 percent improvement in grammaticality over Centrum (from 3.57 to 4.56). Furthermore, referential clarity improves by 24.0 percent (from 3.67 to 4.55), while structure coherence gains 32.6 percent (from 3.37 to 4.47).}

\textcolor{black}{Baseline performance reveals important quality gaps. PRIMERA demonstrates the weakest results, scoring only 2.75 in grammaticality and structure coherence. This indicates significant limitations in language generation quality for supervised approaches. Therefore, MoA's multi-agent coordination produces summaries with superior fluency and logical organization compared to traditional supervised methods.}

\textcolor{black}{\begin{table}[ht]
\centering
\caption{\textcolor{black}{Human evaluation results on VN-MDS dataset (Vietnamese)}}
\label{tab:human_eval_vietnamese}
\begin{tabular*}{\columnwidth}{@{\extracolsep{\fill}}lcccc@{}}
\toprule
\textcolor{black}{\textbf{Model}} & \textcolor{black}{\textbf{Gram.}} & \textcolor{black}{\textbf{Non-red.}} & \textcolor{black}{\textbf{Ref. Clar.}} & \textcolor{black}{\textbf{Str. \& Coh.}} \\
\midrule
\textcolor{black}{\textbf{MoA (Ours)}} & \textcolor{black}{\textbf{4.20}} & \textcolor{black}{\textbf{4.34}} & \textcolor{black}{\textbf{4.29}} & \textcolor{black}{\textbf{4.09}} \\
\textcolor{black}{BertVBD} & \textcolor{black}{3.93} & \textcolor{black}{4.20} & \textcolor{black}{4.02} & \textcolor{black}{3.69} \\
\bottomrule
\end{tabular*}
\end{table}}

\textcolor{black}{Table~\ref{tab:human_eval_vietnamese} presents human evaluation results on VN-MDS dataset. We compare MoA against BertVBD, a competitive training-free Vietnamese baseline. MoA achieves consistent superiority across all evaluation dimensions, validating language-agnostic effectiveness.}

\textcolor{black}{MoA demonstrates strong performance on Vietnamese summarization tasks. Specifically, MoA achieves 4.20 in grammaticality, 4.34 in non-redundancy, 4.29 in referential clarity, and 4.09 in structure coherence. These scores outperform BertVBD across all dimensions. In particular, structure coherence improves by 10.8 percent (from 3.69 to 4.09). Moreover, grammaticality gains 6.9 percent (from 3.93 to 4.20), while referential clarity increases by 6.7 percent (from 4.02 to 4.29). Furthermore, non-redundancy improves by 3.3 percent (from 4.20 to 4.34).}

\textcolor{black}{Cross-language validation confirms architectural robustness. The consistent pattern of MoA superiority across both English and Vietnamese datasets demonstrates language-agnostic effectiveness. MoA maintains strong performance (4.09 to 4.67 range) across diverse linguistic contexts without language-specific tuning. Therefore, human evaluation results complement automatic metrics, providing convergent evidence that MoA produces summaries with superior fluency, coherence, and overall quality from human perspective.}


\section{Conclusions and Future Work}\label{sec:conclusion}

\textcolor{black}{We present MoA, a novel architectural framework for training-free multi-document summarization that coordinates three specialized agents. Our approach delivers three key innovations that advance the field. First, the KGSum agent constructs knowledge graphs to model inter-document relationships and contradictions explicitly. Second, the AMF mechanism fuses agent outputs using metadata-driven strategies. Third, the framework operates without task-specific training while maintaining systematic coordination for complex multi-document scenarios. Furthermore, experimental validation across diverse datasets demonstrates consistent performance gains over both training-free and supervised baselines, establishing the effectiveness of our architectural approach in learning complex inter-document relationships. Moreover, human evaluation confirms superior summary quality improvements in coherence, informativeness, and conciseness.}

\textcolor{black}{Our framework successfully leverages existing language models while providing architectural innovations that enhance their capability to handle complex multi-document relationships. Furthermore, evaluation on established benchmarks provides solid validation for our architectural approach. Future research will extend this framework to specialized domains including technical, legal, and creative content where complex inter-document relationships present greater challenges.} \textcolor{red}{We will also examine how different backbone LLMs influence the relative contribution of each agent within multi-agent summarization frameworks, including our proposed MoA framework.}

\bibliographystyle{plainnat}
\bibliography{sn-bibliography}

\newpage


\begin{appendices}
\section{Prompt Templates for MoA Framework}\label{appendix:kgsum_prompts}
{\color{black}
This appendix presents all prompt templates used throughout the MoA framework. These include KGSum Agent prompts for entity extraction (\(\Pi_e\)), relation classification (\(\Pi_{\text{rel}}\)), community summarization (\(\Pi_s\)), and meta-aware fusion (\(\Pi_f\)), along with Abstractor Agent prompts for initial generation (\(\Pi_{\text{init}}\)) and iterative refinement (\(\Pi_{\text{ref}}\)), and baseline few-shot prompts (\(\Pi_{\text{few}}\)) for comparison.

\begin{figure}[htbp]
    \centering
    \includegraphics[width=0.9\textwidth]{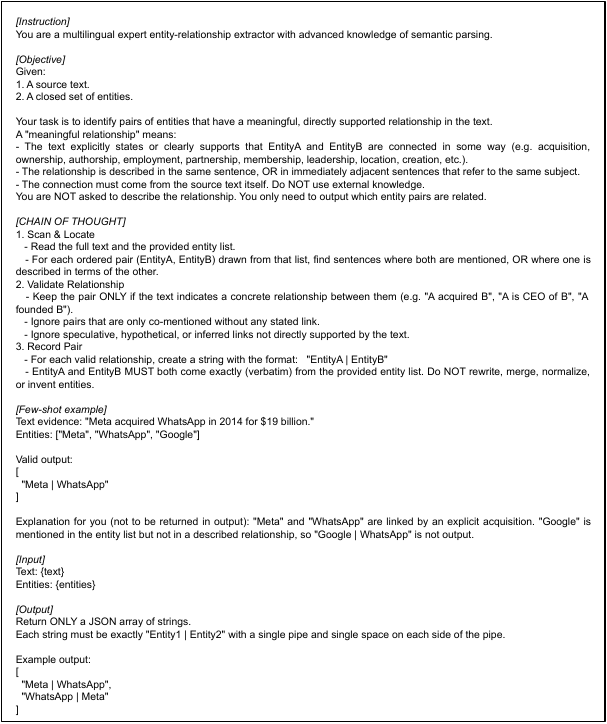}
    \caption{\textcolor{black}{Entity Extraction Prompt (\(\Pi_e\)) used by the KGSum Agent to identify relationships between entities.}}
    \label{fig:entity_extraction_prompt}
\end{figure}

\begin{figure}[htbp]
    \centering
    \includegraphics[width=0.9\textwidth]{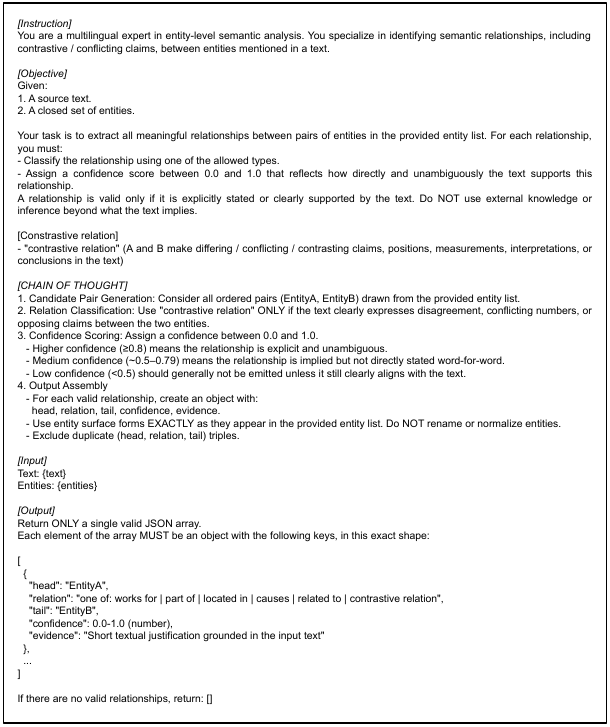}
    \caption{\textcolor{black}{Relation and Contrast Extraction Prompt (\(\Pi_{\text{rel}}\)) used by the KGSum Agent to classify relationships between entity pairs, including contrastive relations.}}
    \label{fig:relation_extraction_prompt}
\end{figure}

\begin{figure}[htbp]
    \centering
    \includegraphics[width=0.9\textwidth]{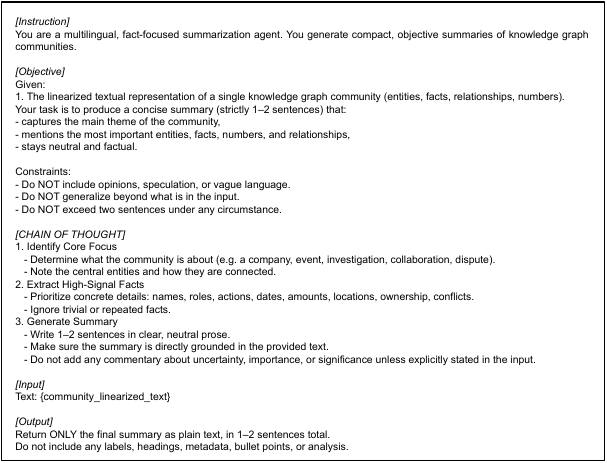}
    \caption{\textcolor{black}{Community Summary Generation Prompt (\(\Pi_s\)) used by the KGSum Agent to produce concise summaries for individual KG communities.}}
    \label{fig:community_summary_prompt}
\end{figure}

\begin{figure}[htbp]
    \centering
    \includegraphics[width=0.9\textwidth]{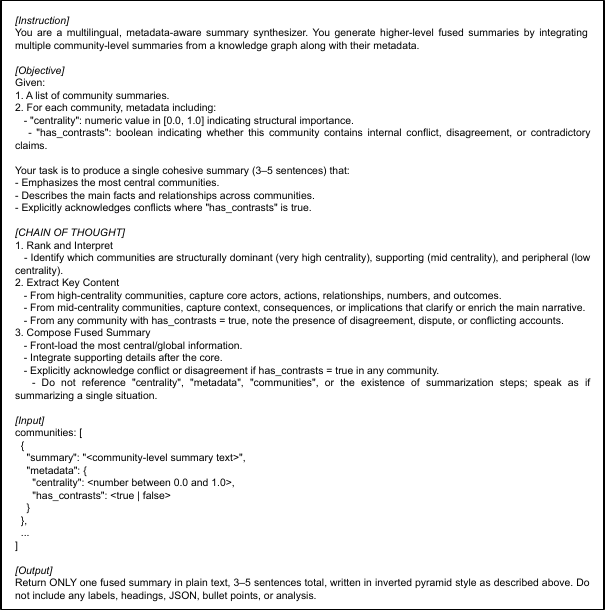}
    \caption{\textcolor{black}{Meta-Aware Fusion Prompt (\(\Pi_f\)) used by the KGSum Agent to synthesize a final summary from community summaries, guided by metadata.}}
    \label{fig:meta_aware_fusion_prompt}
\end{figure}

\begin{figure}[htbp]
    \centering
    \includegraphics[width=0.9\textwidth]{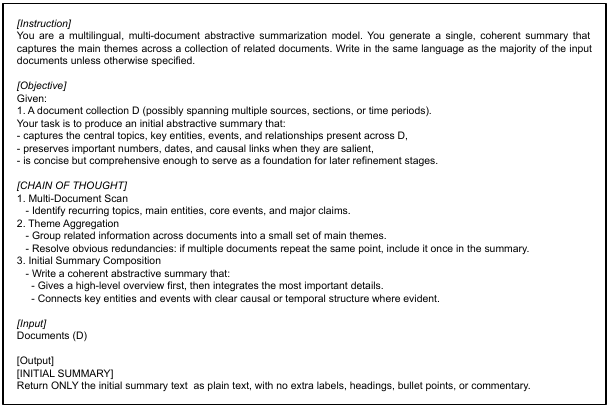}
    \caption{\textcolor{black}{Abstractor Agent Prompt Template for Initial Summary Generation (\(\Pi_{\text{init}}\)), guiding the LLM to produce a first-pass abstractive summary from the document set.}}
    \label{fig:abstractor_init_prompt}
\end{figure}

\begin{figure}[htbp]
    \centering
    \includegraphics[width=0.9\textwidth]{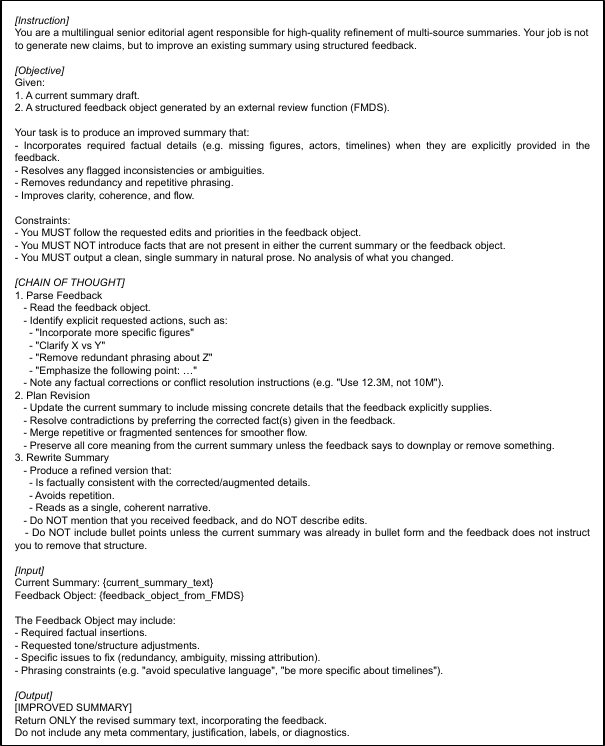}
    \caption{\textcolor{black}{Abstractor Agent Prompt Template for Iterative Refinement (\(\Pi_{\text{ref}}\)), guiding the LLM to improve summaries based on structured feedback.}}
    \label{fig:abstractor_agent_prompt}
\end{figure}

\begin{figure}
    \centering
    \includegraphics[width=0.9\textwidth]{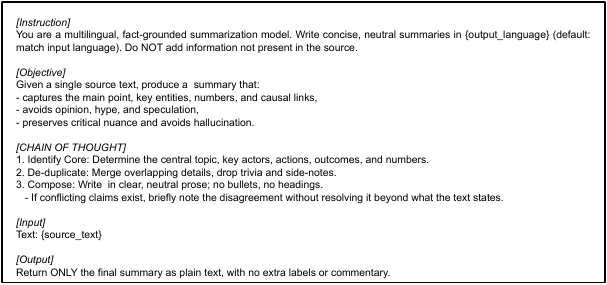}
    \caption{\textcolor{black}{Few-Shot Prompt Template (\(\Pi_{\text{few}}\)) used to compare between performance of LLMs with our MoA framework and standalone LLM summarization.}}
\end{figure}
}

\textcolor{black}{\section{Detailed Analysis of KGSum Agent}\label{appendix:kgsum_detailed_analysis}}

{\color{black}

This section provides comprehensive analysis of the KGSum Agent's key design decisions. We validate the effectiveness of the Leiden algorithm for community detection and examine the impact of contrastive relation extraction on summary quality. These experiments demonstrate that both architectural choices contribute substantially to the agent's performance.

\subsection{Leiden Algorithm Performance Validation}\label{appendix:leiden_eval}

We validate the Leiden algorithm choice through comparative evaluation against Standard Louvain and a no-community-detection baseline on the Multi-News dataset. Table~\ref{tab:leiden_comparison} presents results for both structural quality (Modularity) and downstream summarization performance (ROUGE scores).

\begin{table}[ht]
\centering
\caption{\textcolor{black}{Comparison of community detection algorithms on Multi-News}}
\label{tab:leiden_comparison}
\begin{tabular*}{\columnwidth}{@{\extracolsep{\fill}}lcccc@{}}
\toprule
\textbf{Configuration} & \textbf{Modularity} & \textbf{ROUGE-1} & \textbf{ROUGE-2} & \textbf{ROUGE-L} \\
\midrule
No Community (Baseline) & 0.0000 & 0.4955 & 0.2183 & 0.2241 \\
Standard Louvain & 0.5470 & 0.5281 & 0.2314 & 0.2655 \\
\textbf{Leiden (Our Approach)} & \textbf{0.5512} & \textbf{0.5497} & \textbf{0.2490} & \textbf{0.2940} \\
\bottomrule
\end{tabular*}
\end{table}

Leiden achieves superior modularity (0.5512 vs 0.5470 for Louvain), indicating more robust community structure. Moreover, Leiden translates this structural advantage into downstream performance gains of 4.1 percent in ROUGE-1, 7.6 percent in ROUGE-2, and 8.8 percent in ROUGE-L over Standard Louvain. Therefore, the quality of community detection directly affects summarization effectiveness. By producing thematically coherent communities, Leiden enables meta-aware prompts to operate more effectively, improving content selection and semantic coherence.

\subsection{Impact of Contrastive Relation Extraction}\label{appendix:contrastive_analysis}

To demonstrate the importance of contrastive relation extraction in the KGSum Agent, we conduct isolated ablation experiments comparing performance with and without contrastive relations on Multi-News and VN-MDS datasets using only the standalone KGSum agent. We evaluate two configurations: Without Contrastive extracts only standard relations (e.g., ``related to'', ``part of''), while With Contrastive additionally identifies contrastive relations (e.g., ``contradicts'', ``differs from'') using prompt $\Pi_{\text{rel}}$ (Appendix~\ref{appendix:kgsum_prompts}).

Table~\ref{tab:contrastive_comparison} presents performance comparison. The With Contrastive configuration achieves substantial gains on Multi-News: 15.6 percent improvement in ROUGE-1, 100.0 percent in ROUGE-2, and 25.0 percent in ROUGE-L. On VN-MDS, gains are consistent: 5.0 percent in ROUGE-1, 5.2 percent in ROUGE-2, and 5.1 percent in ROUGE-L. The larger gains on Multi-News suggest that contrastive information is particularly valuable for English news articles, where multiple sources often present conflicting perspectives.

\begin{table}[ht]
\centering
\caption{\textcolor{black}{Effect of contrastive relation extraction on standalone KGSum}}
\label{tab:contrastive_comparison}
\begin{tabular*}{\columnwidth}{@{\extracolsep{\fill}}lcccccc@{}}
\toprule
& \multicolumn{3}{c}{\textcolor{black}{\textbf{Multi-News}}} & \multicolumn{3}{c}{\textcolor{black}{\textbf{VN-MDS}}} \\
\cmidrule(lr){2-4}\cmidrule(lr){5-7}
\textcolor{black}{\textbf{Configuration}} & \textcolor{black}{\textbf{R1}} & \textcolor{black}{\textbf{R2}} & \textcolor{black}{\textbf{RL}} & \textcolor{black}{\textbf{R1}} & \textcolor{black}{\textbf{R2}} & \textcolor{black}{\textbf{RL}} \\
\midrule
\textcolor{black}{Without Contrastive} & \textcolor{black}{0.32} & \textcolor{black}{0.04} & \textcolor{black}{0.12} & \textcolor{black}{0.5689} & \textcolor{black}{0.0873} & \textcolor{black}{0.2494} \\
\textcolor{black}{\textbf{With Contrastive}} & \textcolor{black}{\textbf{0.37}} & \textcolor{black}{\textbf{0.08}} & \textcolor{black}{\textbf{0.15}} & \textcolor{black}{\textbf{0.5976}} & \textcolor{black}{\textbf{0.0918}} & \textcolor{black}{\textbf{0.2622}} \\
\bottomrule
\end{tabular*}
\end{table}

To illustrate how contrastive relations improve quality, we examine Sample 3 from Multi-News discussing Microsoft's Nokia acquisition. Figure~\ref{fig:contrastive_example} compares summaries with and without contrastive extraction. The contrastive-aware summary demonstrates three critical improvements. \textbf{First}, it explicitly contrasts Nokia's historical dominance (35 percent market share) with its decline, providing context for the deal's strategic necessity. \textbf{Second}, it frames Microsoft's 15 percent target as ``lackluster'' relative to Apple and Google, capturing analyst skepticism. \textbf{Third}, it emphasizes the fundamental tension between Microsoft's hardware ambitions and the shifting mobile ecosystem, rather than simply listing concerns. Therefore, contrastive relations enable the KGSum agent to capture conflicting perspectives, temporal contrasts, and competitive tensions that would otherwise be lost, explaining the substantial ROUGE improvements in Table~\ref{tab:contrastive_comparison}.

\begin{figure}[htbp]
    \centering
	\includegraphics[width=0.9\textwidth]{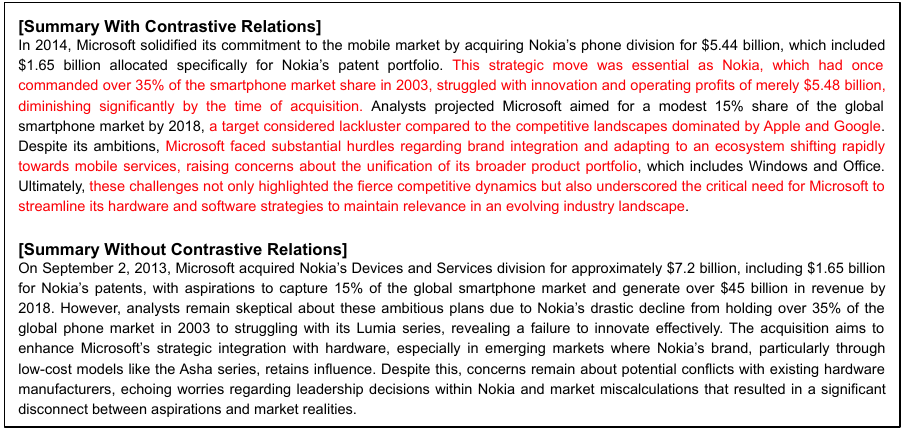}
	\caption{\textcolor{black}{Comparison of summaries generated by KGSum Agent without and with contrastive relation extraction for Sample 3 from Multi-News. The contrastive-aware summary captures nuanced tensions and conflicting perspectives. Key improvements are highlighted in red.}}
    \label{fig:contrastive_example}
\end{figure}

\subsection{Error Propagation Analysis in Hybrid Extraction}\label{appendix:error_propagation}

The KGSum agent employs a hybrid extraction approach combining spaCy for entity recognition, rule-based filtering, and LLM reasoning for relation classification. To address concerns about error propagation from noisy KG components through community detection to final summaries, we conduct controlled noise injection experiments on the Multi-News dataset. We evaluate three scenarios: Clean Baseline (0 percent entity/relation noise), Moderate Noise (20 percent entity, 30 percent relation), and High Noise (40 percent entity, 50 percent relation). Noise is randomly injected as spurious entities and incorrect relations before community detection. We measure KG Noise Score (proportion of injected noise retained), Noise Amplification (noise ratio in summaries to input), and ROUGE performance.

Table~\ref{tab:noise_robustness} presents results across all scenarios. The hybrid approach demonstrates strong robustness. KG Noise Score remains 0.0000 across all conditions, indicating the hybrid filtering successfully rejects injected noise before it enters the graph structure. The combination of spaCy statistical recognition, rule-based validation, and LLM reasoning provides multiple verification layers. Moreover, Noise Amplification is consistently 0.0000, demonstrating that the multi-stage architecture prevents error cascading even if minimal noise enters. ROUGE scores show graceful degradation: ROUGE-1 decreases from 37.0 (Clean) to 36.8 (Moderate) to 36.5 (High), ROUGE-2 from 4.0 to 3.9 to 3.8, and ROUGE-L from 12.0 to 11.9 to 11.7. The minimal performance degradation (less than 2 percent across all metrics) despite substantial injected noise (up to 40 percent entity and 50 percent relation noise) validates the robustness of the hybrid extraction architecture. Therefore, the hybrid extraction architecture acts as a robust filtering mechanism, preventing noisy components from contaminating downstream processes and ensuring reliable performance for real-world deployment.

\begin{table}[ht]
\centering
\caption{\textcolor{black}{Robustness of KGSum under controlled noise on Multi-News}}
\label{tab:noise_robustness}
\begin{tabular*}{\columnwidth}{@{\extracolsep{\fill}}lcccccc@{}}
\toprule
\textcolor{black}{\textbf{Scenario}} & \textcolor{black}{\textbf{KG Noise}} & \textcolor{black}{\textbf{Amplification}} & \textcolor{black}{\textbf{R1}} & \textcolor{black}{\textbf{R2}} & \textcolor{black}{\textbf{RL}} \\
\midrule
\textcolor{black}{Clean (0\% E, 0\% R)} & \textcolor{black}{0.0000} & \textcolor{black}{0.0000} & \textcolor{black}{37.0 ($\pm$1.2)} & \textcolor{black}{4.0} & \textcolor{black}{12.0} \\
\textcolor{black}{Moderate (20\% E, 30\% R)} & \textcolor{black}{0.0000} & \textcolor{black}{0.0000} & \textcolor{black}{36.8 ($\pm$1.3)} & \textcolor{black}{3.9} & \textcolor{black}{11.9} \\
\textcolor{black}{High (40\% E, 50\% R)} & \textcolor{black}{0.0000} & \textcolor{black}{0.0000} & \textcolor{black}{36.5 ($\pm$1.4)} & \textcolor{black}{3.8} & \textcolor{black}{11.7} \\
\bottomrule
\end{tabular*}
\vspace{2mm}
\centering
\footnotesize \textcolor{black}{Note: E = Entity noise, R = Relation noise. Standard deviations shown for R1.}
\end{table}
}

\textcolor{black}{\section{Detailed Dataset Descriptions}\label{appendix:dataset_details}}

\textcolor{black}{\subsection{Dataset Overview}\label{appendix:dataset_overview}}

\textcolor{black}{We evaluate MoA on four diverse MDS datasets spanning multiple domains and languages. Multi-News \citep{fabbri_multi-news_2019} consists of news articles from multiple sources covering the same event, requiring synthesis of potentially conflicting perspectives and redundant information. Multi-XScience \citep{lu_multi-xscience_2020} comprises scientific papers with related work sections, requiring capture of technical terminology, citation relationships, and methodological comparisons. VN-MDS\footnote{https://github.com/lupanh/VietnameseMDS} contains Vietnamese news articles from multiple outlets, testing cross-lingual applicability and cultural context handling. ViMs \citep{tran2020vims} includes diverse Vietnamese documents spanning multiple genres, evaluating robustness to varied writing styles and topical domains. This evaluation demonstrates our framework's capability to handle complex inter-document relationships across diverse domains and languages.}

\textcolor{black}{For all datasets, we apply consistent preprocessing to ensure clean LLM input. Input documents are segmented into sentences using language-specific tokenizers: spaCy for English datasets and Vietnamese-appropriate tokenizers for Vietnamese datasets. Special characters including excessive punctuation, non-standard symbols, and formatting artifacts are removed to prevent tokenization errors. The training set is used solely for prompt optimization, as MoA is a training-free framework. Test splits are reserved for evaluation to ensure fair comparison with published baselines. Therefore, all reported results reflect true generalization performance on held-out test data.}

\textcolor{black}{\subsection{Comparative Complexity Analysis}\label{appendix:dataset_complexity}}

\textcolor{black}{To understand performance differences between English and Vietnamese datasets, we conduct systematic complexity analysis. Vietnamese test sets contain substantially fewer samples (29 to 45) compared to English datasets (5,000 plus), making direct comparison challenging. Therefore, we sample 100 instances from training sets of all four datasets to enable fair cross-dataset complexity comparison. We evaluate three key dimensions: Type-Token Ratio or TTR (lexical diversity, higher is more complex), N-gram Overlap (content redundancy, lower is more complex), and Extractiveness (proportion of summary directly copied from source, lower is more complex). Table~\ref{tab:dataset_complexity_comparison} presents comprehensive metrics. \textcolor{black}{\textbf{Bold red}} indicates highest complexity, \textbf{bold black} indicates second-highest complexity within each metric category.}

\begin{table}[ht]
\centering
\caption{\textcolor{black}{Comparative dataset complexity metrics}}
\label{tab:dataset_complexity_comparison}
\small
\begin{tabular*}{\columnwidth}{@{\extracolsep{\fill}}lcccc@{}}
\toprule
\textbf{Complexity Metric} & \textbf{Multi-News} & \textbf{Multi-XScience} & \textbf{VN-MDS} & \textbf{ViMs} \\
\midrule
\multicolumn{5}{l}{\textit{Lexical Diversity (higher = more challenging)}} \\
Document TTR & \textcolor{black}{\textbf{0.1135}} & \textbf{0.1106} & 0.0314 & 0.0200 \\
Summary TTR & \textcolor{black}{\textbf{0.2339}} & \textbf{0.2246} & 0.1204 & 0.1081 \\
\midrule
\multicolumn{5}{l}{\textit{Content Redundancy (lower = more challenging)}} \\
Bigram overlap & \textbf{0.0445} & \textcolor{black}{\textbf{0.0273}} & 0.1253 & 0.1736 \\
Trigram overlap & \textbf{0.0174} & \textcolor{black}{\textbf{0.0083}} & 0.0772 & 0.1113 \\
\midrule
\multicolumn{5}{l}{\textit{Summary Characteristics (lower extractiveness = more challenging)}} \\
Avg. extractiveness & \textbf{0.2760} & \textcolor{black}{\textbf{0.1075}} & 0.9267 & 0.9601 \\
Avg. novel content & \textbf{0.7240} & \textcolor{black}{\textbf{0.8925}} & 0.0733 & 0.0399 \\
\bottomrule
\end{tabular*}
\end{table}

\textcolor{black}{The analysis reveals systematic complexity differences strongly favoring English datasets. English datasets demonstrate substantially higher lexical diversity with document TTR 4.4 times higher than Vietnamese (0.1121 versus 0.0257 average), requiring models to handle diverse linguistic expressions and paraphrasing capabilities. Vietnamese datasets exhibit significantly higher content redundancy with bigram overlap 4.2 times higher (0.1495 versus 0.0359 average) and trigram overlap 7.4 times higher (0.0942 versus 0.0128 average), simplifying information synthesis through multiple overlapping formulations. Vietnamese summaries are highly extractive (94.3 percent versus 19.2 percent for English), inflating ROUGE scores through trivial exact n-gram matching rather than demanding abstractive synthesis. Therefore, higher absolute ROUGE scores on Vietnamese datasets reflect inherently less challenging evaluation conditions rather than superior model capabilities for Vietnamese language processing.}

\textcolor{black}{\section{Hyperparameter Configuration}\label{appendix:hyperparams}}

\textcolor{black}{Table~\ref{tab:hyperparams} presents complete hyperparameter configuration for the MoA framework. All hyperparameter values are determined through systematic prompt optimization on the training set, balancing performance quality with computational efficiency. The Extractor agent employs sentence scoring weight $\lambda = 0.7$ to balance lexical and positional importance, with top-k sentence proportion $\alpha$ targeting approximately 15 percent of source length for sufficient context while maintaining computational efficiency. Redundancy threshold $\theta_r = 0.9$ filters near-duplicate sentences to reduce information redundancy. The KGSum agent uses entity similarity threshold $\theta_s = 0.85$ for entity resolution, relation confidence threshold $\tau_r = 0.6$ for relation extraction quality, and selects $k = 5$ top communities with minimum size 3 and maximum cluster size 50 to balance coverage with computational tractability. The Leiden resolution parameter is set to 1.0 for standard modularity optimization. The Abstractor agent performs up to $T = 2$ refinement iterations with quality threshold $\tau_q^{\text{Abs}} = 0.7$ for triggering refinement, balancing quality improvement with API cost efficiency.}

\begin{table}[ht]
\centering
\caption{\textcolor{black}{Hyperparameter configuration}}
\label{tab:hyperparams}
\begin{tabular}{@{}lll@{}}
\toprule
\textbf{Parameter} & \textbf{Component} & \textbf{Value} \\
\midrule
\multicolumn{3}{l}{\textit{MoA framework hyperparameters}} \\
Sentence Scoring Weight ($\lambda$) & Extractor agent & 0.7 \\
Top-k Sentences Proportion ($\alpha$) & Extractor agent & Target: $\sim$15\% of source length \\
Redundancy Threshold ($\theta_r$) & Extractor agent & 0.9 \\
Entity Similarity Threshold ($\theta_s$) & KGSum agent & 0.85 \\
Relation Confidence Threshold ($\tau_r$) & KGSum agent & 0.6 \\
Number of Top Communities ($k$) & KGSum agent & 5 \\
Minimum Community Size & KGSum agent & 3 \\
Maximum Cluster Size & KGSum agent & 50 \\
Leiden Resolution Parameter & KGSum agent & 1.0 \\
Max Refinement Iterations ($T$) & Abstractor agent & 2 \\
Quality for Refinement ($\tau_q^{\text{Abs}}$) & Abstractor agent & 0.7 \\
\midrule
\multicolumn{3}{l}{\textit{LLM configuration}} \\
LLM Top-p & All agents & 0.8 \\
\midrule
\multicolumn{3}{l}{\textit{Computational Resources}} \\
GPU & Experimentation & NVIDIA Quadro RTX 8000 \\
Number of Runs & Statistical Validation & 3 independent runs (averaged results) \\
\bottomrule
\end{tabular}
\small
\textcolor{black}{\textit{Note: Prompt design strategies are detailed in Appendix~\ref{appendix:kgsum_prompts}.}}
\end{table}

\textcolor{black}{We set LLM top-p sampling to 0.8 for all agents to balance output diversity with response stability. This value ensures sufficient creativity for summarization while maintaining consistency across multiple runs. All experiments are conducted on NVIDIA Quadro RTX 8000 GPU with sufficient memory for batch processing and caching. We perform three independent experimental runs for statistical validation, reporting mean and standard deviation to quantify performance variability and ensure reproducibility. Therefore, all reported results reflect averaged performance across multiple runs, demonstrating the stability and reliability of our framework across different random seeds and initialization conditions.}

\textcolor{black}{\section{API Cost Analysis for Closed-Source Models}\label{appendix:cost_analysis}}

\textcolor{black}{To address practical deployment considerations, we analyze API costs for closed-source LLMs across all evaluation datasets. Costs represent actual expenditures from a single experimental run per dataset, calculated based on official API pricing structures from respective providers (OpenAI, Google, xAI). All costs are reported in USD and include regular input tokens, cached input tokens, and output tokens. This analysis enables informed model selection decisions for real-world applications balancing performance and budget constraints.}

\textcolor{black}{Table~\ref{tab:api_costs} presents comprehensive API cost breakdown across four closed-source models. The analysis reveals substantial cost variations. GPT-4.1-mini achieves superior performance but incurs the highest costs across all datasets: \$94.55 on Multi-News (\$0.0168 per sample), \$58.39 on Multi-XScience (\$0.0115 per sample), \$0.20 on VN-MDS (\$0.0070 per sample), and \$0.51 on ViMs (\$0.0113 per sample). In contrast, Gemini-1.5-flash-8b offers the most cost-effective solution with costs ranging from \$0.02 on VN-MDS to \$10.07 on Multi-News, representing 5.6 to 10.6 times lower costs than GPT-4.1-mini while maintaining competitive performance. GPT-4o-mini and Grok-3-mini provide intermediate cost-performance tradeoffs. Moreover, input token caching reduces costs by 4.4 to 6.3 percent across models, demonstrating efficiency gains from prompt reuse in multi-document processing. Therefore, model selection should consider the performance-cost tradeoff based on application requirements and budget constraints.}

\begin{table*}[htbp]
\centering
\small
\setlength{\tabcolsep}{6pt}
\caption{\textcolor{black}{API cost analysis for closed-source LLMs}}
\label{tab:api_costs}
\begin{tabular*}{\columnwidth}{@{\extracolsep{\fill}}lccccc@{}}
\toprule
\textcolor{black}{Model} & \textcolor{black}{Reg. In (\$)} & \textcolor{black}{Cache In (\$)} & \textcolor{black}{Out (\$)} & \textcolor{black}{Total (\$)} & \textcolor{black}{Per Sample (\$)} \\
\midrule
\multicolumn{6}{c}{\textcolor{black}{Multi-News (5,622 samples)}} \\
\textcolor{black}{GPT-4.1-mini} & \textcolor{black}{68.42} & \textcolor{black}{4.28} & \textcolor{black}{21.86} & \textcolor{black}{94.55} & \textcolor{black}{0.0168} \\
\textcolor{black}{GPT-4o-mini} & \textcolor{black}{25.66} & \textcolor{black}{3.21} & \textcolor{black}{8.20} & \textcolor{black}{37.06} & \textcolor{black}{0.0066} \\
\textcolor{black}{Gemini-1.5-flash-8b} & \textcolor{black}{6.41} & \textcolor{black}{1.60} & \textcolor{black}{2.05} & \textcolor{black}{10.07} & \textcolor{black}{0.0018} \\
\textcolor{black}{Grok-3-mini} & \textcolor{black}{42.76} & \textcolor{black}{3.21} & \textcolor{black}{6.83} & \textcolor{black}{52.80} & \textcolor{black}{0.0094} \\
\midrule
\multicolumn{6}{c}{\textcolor{black}{Multi-XScience (5,093 samples)}} \\
\textcolor{black}{GPT-4.1-mini} & \textcolor{black}{36.32} & \textcolor{black}{2.27} & \textcolor{black}{19.80} & \textcolor{black}{58.39} & \textcolor{black}{0.0115} \\
\textcolor{black}{GPT-4o-mini} & \textcolor{black}{13.62} & \textcolor{black}{1.70} & \textcolor{black}{7.43} & \textcolor{black}{22.75} & \textcolor{black}{0.0045} \\
\textcolor{black}{Gemini-1.5-flash-8b} & \textcolor{black}{3.40} & \textcolor{black}{0.85} & \textcolor{black}{1.86} & \textcolor{black}{6.11} & \textcolor{black}{0.0012} \\
\textcolor{black}{Grok-3-mini} & \textcolor{black}{22.70} & \textcolor{black}{1.70} & \textcolor{black}{6.19} & \textcolor{black}{30.59} & \textcolor{black}{0.0060} \\
\midrule
\multicolumn{6}{c}{\textcolor{black}{VN-MDS (29 samples)}} \\
\textcolor{black}{GPT-4.1-mini} & \textcolor{black}{0.09} & \textcolor{black}{0.01} & \textcolor{black}{0.11} & \textcolor{black}{0.20} & \textcolor{black}{0.0070} \\
\textcolor{black}{GPT-4o-mini} & \textcolor{black}{0.03} & \textcolor{black}{0.00} & \textcolor{black}{0.04} & \textcolor{black}{0.08} & \textcolor{black}{0.0027} \\
\textcolor{black}{Gemini-1.5-flash-8b} & \textcolor{black}{0.01} & \textcolor{black}{0.00} & \textcolor{black}{0.01} & \textcolor{black}{0.02} & \textcolor{black}{0.0007} \\
\textcolor{black}{Grok-3-mini} & \textcolor{black}{0.05} & \textcolor{black}{0.00} & \textcolor{black}{0.04} & \textcolor{black}{0.09} & \textcolor{black}{0.0032} \\
\midrule
\multicolumn{6}{c}{\textcolor{black}{ViMs (45 samples)}} \\
\textcolor{black}{GPT-4.1-mini} & \textcolor{black}{0.31} & \textcolor{black}{0.02} & \textcolor{black}{0.17} & \textcolor{black}{0.51} & \textcolor{black}{0.0113} \\
\textcolor{black}{GPT-4o-mini} & \textcolor{black}{0.12} & \textcolor{black}{0.01} & \textcolor{black}{0.07} & \textcolor{black}{0.20} & \textcolor{black}{0.0044} \\
\textcolor{black}{Gemini-1.5-flash-8b} & \textcolor{black}{0.03} & \textcolor{black}{0.01} & \textcolor{black}{0.02} & \textcolor{black}{0.05} & \textcolor{black}{0.0012} \\
\textcolor{black}{Grok-3-mini} & \textcolor{black}{0.19} & \textcolor{black}{0.01} & \textcolor{black}{0.05} & \textcolor{black}{0.26} & \textcolor{black}{0.0059} \\
\bottomrule
\end{tabular*}
\vspace{2mm}

\footnotesize Note: Reg. In = Regular Input tokens, Cache In = Cached Input tokens, Out = Output tokens. \textbf{Bold} indicates highest cost within each dataset.
\end{table*}

\textcolor{black}{\section{Statistical Significance Validation}\label{appendix:statistical_validation}}

\textcolor{black}{To validate the reliability of our performance improvements, we conduct dataset-level statistical validation using the sign test. This non-parametric method evaluates whether MoA systematically outperforms published baselines across 12 independent dataset-metric combinations derived from 4 evaluation datasets (Multi-News, Multi-XScience, VN-MDS, ViMs) and 3 ROUGE metrics (R1, R2, RL). For each combination, we compare MoA against the best published baseline as reported in Table~\ref{tab:english_summary_results} and Table~\ref{tab:vietnamese_summary_results}. The sign test counts comparisons where MoA wins. Under the null hypothesis of equal performance, the expected win proportion is 0.5. Therefore, observing significantly higher win rates provides evidence of systematic superiority.}

\textcolor{black}{Table~\ref{tab:statistical_validation} presents validation results. MoA achieves superior performance in all 12 dataset-metric combinations (100 percent win rate). The one-tailed binomial test yields $p < 0.001$ for observing 12 wins out of 12 trials, providing conclusive evidence that MoA's superior performance reflects genuine systematic improvements rather than random chance. Moreover, the consistency of wins across diverse content types (news articles, scientific papers, diverse documents) and languages (English, Vietnamese) demonstrates robustness and generalizability.}

\textcolor{black}{\begin{table}[ht]
\centering
\caption{\textcolor{black}{Statistical validation against best published baselines}}
\label{tab:statistical_validation}
\small
\begin{tabular*}{\columnwidth}{@{\extracolsep{\fill}}llccc@{}}
\toprule
\textcolor{black}{\textbf{Dataset}} & \textcolor{black}{\textbf{Metric}} & \textcolor{black}{\textbf{Best Baseline}} & \textcolor{black}{\textbf{MoA}} & \textcolor{black}{\textbf{Win?}} \\
\midrule
\textcolor{black}{Multi-News} & \textcolor{black}{R1} & \textcolor{black}{0.4570 (Centrum)} & \textcolor{black}{0.5497} & \textcolor{black}{\checkmark} \\
\textcolor{black}{Multi-News} & \textcolor{black}{R2} & \textcolor{black}{0.1680 (Centrum)} & \textcolor{black}{0.2490} & \textcolor{black}{\checkmark} \\
\textcolor{black}{Multi-News} & \textcolor{black}{RL} & \textcolor{black}{0.2320 (Centrum)} & \textcolor{black}{0.2940} & \textcolor{black}{\checkmark} \\
\midrule
\textcolor{black}{Multi-XScience} & \textcolor{black}{R1} & \textcolor{black}{0.3179 (GLIMMER-TTR)} & \textcolor{black}{0.3666} & \textcolor{black}{\checkmark} \\
\textcolor{black}{Multi-XScience} & \textcolor{black}{R2} & \textcolor{black}{0.0481 (GLIMMER-TTR)} & \textcolor{black}{0.0564} & \textcolor{black}{\checkmark} \\
\textcolor{black}{Multi-XScience} & \textcolor{black}{RL} & \textcolor{black}{0.1671 (GLIMMER-TTR)} & \textcolor{black}{0.1973} & \textcolor{black}{\checkmark} \\
\midrule
\textcolor{black}{VN-MDS} & \textcolor{black}{R1} & \textcolor{black}{0.7676 (LatVisNewshead)} & \textcolor{black}{0.8011} & \textcolor{black}{\checkmark} \\
\textcolor{black}{VN-MDS} & \textcolor{black}{R2} & \textcolor{black}{0.5021 (LatVisNewshead)} & \textcolor{black}{0.5232} & \textcolor{black}{\checkmark} \\
\textcolor{black}{VN-MDS} & \textcolor{black}{RL} & \textcolor{black}{0.4739 (LatVisNewshead)} & \textcolor{black}{0.4829} & \textcolor{black}{\checkmark} \\
\midrule
\textcolor{black}{ViMs} & \textcolor{black}{R1} & \textcolor{black}{0.7898 (LatVisNewshead)} & \textcolor{black}{0.8026} & \textcolor{black}{\checkmark} \\
\textcolor{black}{ViMs} & \textcolor{black}{R2} & \textcolor{black}{0.5508 (LatVisNewshead)} & \textcolor{black}{0.5557} & \textcolor{black}{\checkmark} \\
\textcolor{black}{ViMs} & \textcolor{black}{RL} & \textcolor{black}{0.4800 (LatVisNewshead)} & \textcolor{black}{0.4904} & \textcolor{black}{\checkmark} \\
\midrule
\multicolumn{3}{l}{\textcolor{black}{\textbf{Total}}} & \textcolor{black}{\textbf{12/12}} & \textcolor{black}{\textbf{100\%}} \\
\bottomrule
\end{tabular*}
\end{table}}

\textcolor{black}{Beyond statistical significance, we examine practical significance through effect sizes. Improvements range from 4.2 percent (ViMs RL) to 48.2 percent (Multi-News R2), with median improvement of 15.3 percent across all combinations. In particular, substantial gains on challenging English benchmarks (Multi-News, Multi-XScience) demonstrate meaningful performance enhancements for real-world applications. Furthermore, consistent improvements across both training-free and supervised baselines validate that our architectural coordination addresses fundamental limitations in existing MDS methodologies, establishing MoA as a reliable and generalizable solution.}

\textcolor{black}{\section{Human Evaluation Interface and Guidelines}\label{appendix:human_eval_interface}}

\begin{figure}[htbp]
    \centering
    \includegraphics[width=0.9\textwidth]{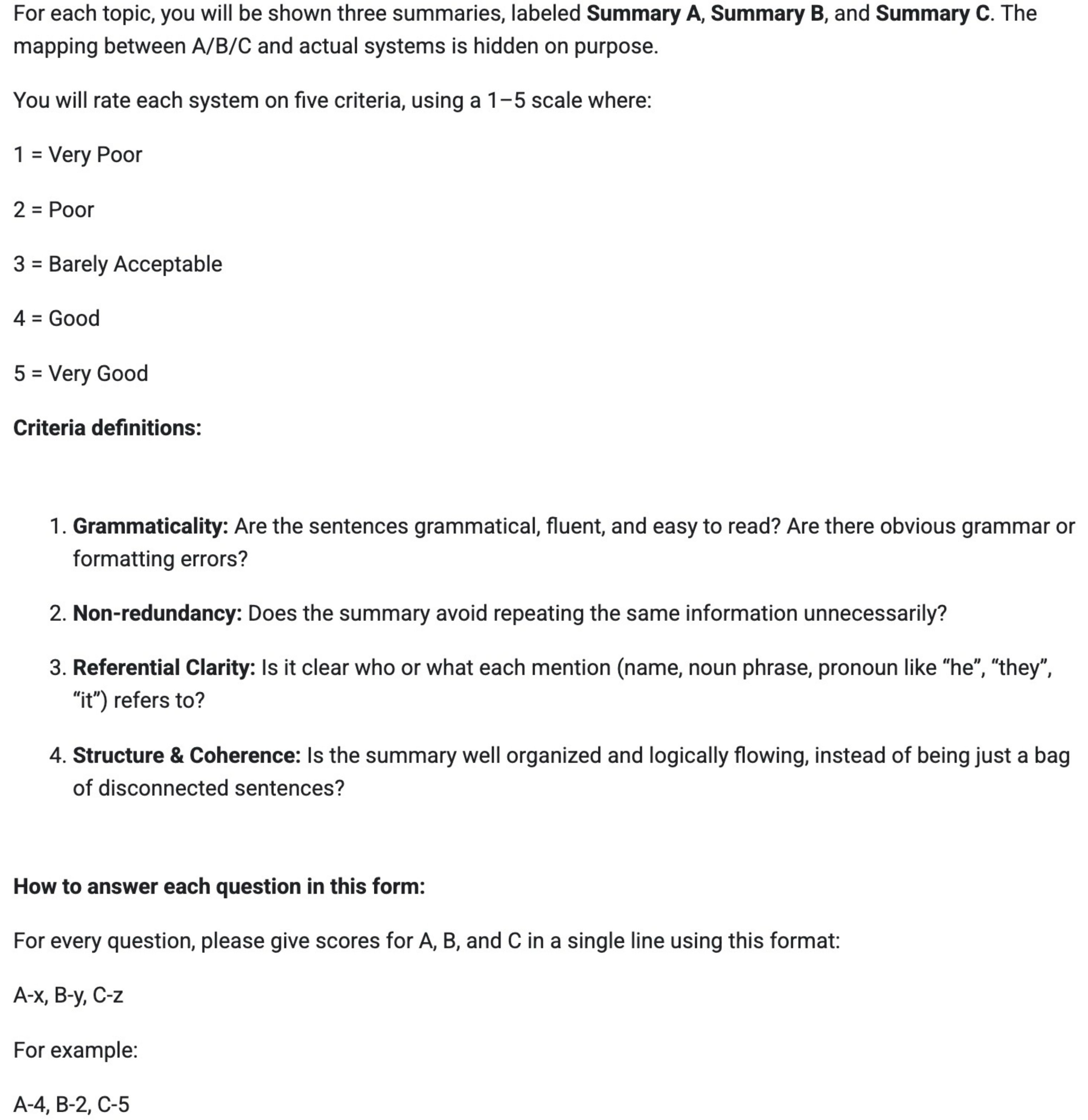}
    \caption{\textcolor{black}{Human Evaluation Interface used for assessing summary quality based on Grammaticality, Non-redundancy, Referential Clarity, and Structure and Coherence criteria.}}
    \label{fig:human_evaluation_interface}
\end{figure}

\textcolor{black}{This appendix describes our human evaluation methodology for assessing summary quality across multiple systems. Figure~\ref{fig:human_evaluation_interface} presents the annotation interface used for comparative evaluation. For each document cluster, participants evaluate three summaries labeled Summary A, Summary B, and Summary C, where the mapping between labels and actual systems is hidden to prevent bias. Participants rate each summary on four quality criteria using a 1 to 5 scale: (1) Grammaticality assesses whether sentences are grammatical, fluent, and easy to read without obvious grammar or formatting errors; (2) Non-redundancy evaluates whether the summary avoids repeating the same information unnecessarily; (3) Referential Clarity examines whether it is clear who or what each mention refers to; (4) Structure and Coherence measures whether the summary is well organized and logically flowing rather than disconnected sentences. The rating scale ranges from 1 (Very Poor) to 5 (Very Good), with intermediate levels at 2 (Poor), 3 (Barely Acceptable), and 4 (Good).}

\textcolor{black}{Participants receive source documents and three anonymized summaries in randomly ordered presentation for each evaluation task. They independently rate each summary on all four criteria after reading source documents to understand the content. The interface requires participants to provide scores for all three summaries (A, B, C) in a single line using the format A-x, B-y, C-z (e.g., A-4, B-2, C-5). Moreover, we implement quality control through comprehensive training materials explaining each criterion with concrete examples, attention check questions to identify inattentive participants, and inter-annotator agreement computation to verify consistency. Participants showing low agreement with others or failing attention checks are excluded from final analysis. Therefore, this rigorous protocol ensures that reported results reflect genuine quality assessments rather than random ratings, providing reliable validation of our framework's performance improvements.}

\textcolor{black}{\section{Impact of Cluster Size on MoA Performance}\label{appendix:scaling_study}}

\textcolor{black}{To evaluate the robustness of the MoA framework under varying document quantities, we conduct a focused scaling study on Multi-News. We systematically compare performance across three cluster size configurations: Small (3 documents per cluster), Medium (5 documents per cluster), and Large (10 documents per cluster). For each configuration, we evaluate 10 randomly sampled clusters with 3 independent runs per sample, reporting mean and standard deviation across 3 runs. All experimental conditions remain constant (GPT-4.1-mini backend, prompt templates, token budget, filtering thresholds, fusion strategies) with only cluster size varying to isolate its effect on framework performance.}

\textcolor{black}{Table~\ref{tab:scaling_multinews} presents scaling results. The framework demonstrates graceful degradation as cluster size increases. Small (3 documents) achieves optimal performance with 60.2 percent ROUGE-1, 28.1 percent ROUGE-2, and 31.8 percent ROUGE-L. Medium (5 documents) shows slight degradation with 55.4 percent ROUGE-1 (4.8 points decrease, 7.97 percent relative), 24.9 percent ROUGE-2 (3.2 points decrease, 11.39 percent relative), and 29.4 percent ROUGE-L (2.4 points decrease, 7.55 percent relative). Large (10 documents) exhibits more substantial degradation with 49.2 percent ROUGE-1 (11.0 points decrease from Small, 18.27 percent relative), 21.8 percent ROUGE-2 (6.3 points decrease, 22.42 percent relative), and 27.2 percent ROUGE-L (4.6 points decrease, 14.47 percent relative).}

\begin{table}[ht]
\centering
\caption{\textcolor{black}{Scaling study on Multi-News by cluster size}}
\label{tab:scaling_multinews}
\begin{tabular*}{\columnwidth}{@{\extracolsep{\fill}}lccc@{}}
\toprule
\textcolor{black}{\textbf{Cluster size}} & \textcolor{black}{\textbf{R1}} & \textcolor{black}{\textbf{R2}} & \textcolor{black}{\textbf{RL}} \\
\midrule
\textcolor{black}{\textbf{Small (3)}}  & \textcolor{black}{\textbf{60.2}$\pm$0.4} & \textcolor{black}{\textbf{28.1}$\pm$0.3} & \textcolor{black}{\textbf{31.8}$\pm$0.3} \\
\textcolor{black}{\textbf{Medium (5)}} & \textcolor{black}{55.4$\pm$0.3} & \textcolor{black}{24.9$\pm$0.3} & \textcolor{black}{29.4$\pm$0.2} \\
\textcolor{black}{\textbf{Large (10)}} & \textcolor{black}{49.2$\pm$0.5} & \textcolor{black}{21.8$\pm$0.4} & \textcolor{black}{27.2$\pm$0.3} \\
\bottomrule
\end{tabular*}
\end{table}

\textcolor{black}{These results demonstrate three critical insights. First, the framework maintains acceptable performance even at 10 documents per cluster, with ROUGE-L degradation limited to 14.47 percent relative to optimal conditions. Second, the degradation pattern is relatively stable from Medium to Large configurations, suggesting architectural robustness beyond typical dataset scales. Third, the performance-cost tradeoff enables informed deployment decisions: applications prioritizing quality can use smaller clusters (3 to 5 documents), while cost-sensitive deployments can scale to larger clusters (10 documents) with graceful degradation. Therefore, the MoA framework demonstrates robust scalability across document quantities, validating its applicability for real-world multi-document summarization scenarios with varying cluster sizes.}

\end{appendices}

\end{document}